\documentclass[10pt,twocolumn,letterpaper]{article}

\usepackage{cvpr}
\usepackage{times}
\usepackage{epsfig}
\usepackage{graphicx}
\usepackage{amsmath}
\usepackage{amssymb}
\usepackage{hhline}
\usepackage{multirow}
\usepackage{array}
\usepackage{tabu}
\usepackage{booktabs}
\usepackage{subfigure}
\usepackage{rotating}

\usepackage[pagebackref=true,breaklinks=true,letterpaper=true,colorlinks,bookmarks=false]{hyperref}

\cvprfinalcopy 

\def\cvprPaperID{1176} 

\ifcvprfinal\fi
\begin{document}

\title{All about Structure: Adapting Structural Information across Domains for Boosting Semantic Segmentation}

\author{
Wei-Lun Chang* \quad Hui-Po Wang* \quad Wen-Hsiao Peng \quad Wei-Chen Chiu\\
National Chiao Tung University, Taiwan\\
{\tt\small \{luckchang.ee06g, a88575847.cs06g, wpeng, walon\}@nctu.edu.tw}
}
\maketitle

\newcommand*\rot{\rotatebox{90}}
\newcolumntype{L}[1]{>{\raggedright\let\newline\\\arraybackslash\hspace{0pt}}m{#1}}
\newcolumntype{C}[1]{>{\centering\let\newline\\\arraybackslash\hspace{0pt}}m{#1}}
\newcolumntype{R}[1]{>{\raggedleft\let\newline\\\arraybackslash\hspace{0pt}}m{#1}}
\newcommand\blfootnote[1]{%
  \begingroup
  \renewcommand\thefootnote{}\footnote{#1}%
  \addtocounter{footnote}{-1}%
  \endgroup
}

\begin{abstract}
In this paper we tackle the problem of unsupervised domain adaptation for the task of semantic segmentation, where we attempt to transfer the knowledge learned upon synthetic datasets with ground-truth labels to real-world images without any annotation. With the hypothesis that the structural content of images is the most informative and decisive factor to semantic segmentation and can be readily shared across domains, we propose a Domain Invariant Structure Extraction (DISE) framework to disentangle images into domain-invariant structure and domain-specific texture representations, which can further realize image-translation across domains and enable label transfer to improve segmentation performance. Extensive experiments verify the effectiveness of our proposed DISE model and demonstrate its superiority over several state-of-the-art approaches. 
\end{abstract}
\blfootnote{*Both authors contribute equally}

\section{Introduction}\label{sec:intro}
Semantic segmentation is to predict pixel-level semantic labels for an image. It is considered one of the most challenging tasks in computer vision. Due to the renaissance of deep learning in recent years, we witness a great leap brought to this task. Since the inception of Fully Convolutional Network (FCN), which is built upon pre-trained classification models (e.g. VGG~\cite{simonyan2014very} and ResNet~\cite{he2016deep}) and deconvolutional layers, numerous techniques have been proposed to advance semantic segmentation, such as enlarging receptive fields~\cite{chen2018deeplab,yu2015multi} and better preserving contextual information~\cite{zhao2017pyramid}, to name a few. However, these approaches rely largely on supervised learning, thereby calling for expensive pixel-level annotations.

To circumvent this issue, one solution is to train segmentation models on synthetic data. The computer graphics technology nowadays is able to synthesize high-quality, photo-realistic images for a virtual scene. It is thus possible to build up a dataset for supervised semantic segmentation (e.g. GTA5 \cite{richter2016playing} and SYNTHIA \cite{ros2016synthia}) based on these synthetic images. During the rendering process, their pixel-level semantic labels are readily available. Nevertheless, segmentation models trained on synthetic datasets often have difficulty achieving satisfactory performance in real-world scenes due to a phenomenon known as \textit{domain shift} -- i.e. synthetic and real-world images can still exhibit considerable difference in their low-level texture appearance.

\begin{figure}[t]
    \centering
    \begin{tabular}{c}
        \includegraphics[width=1\linewidth]{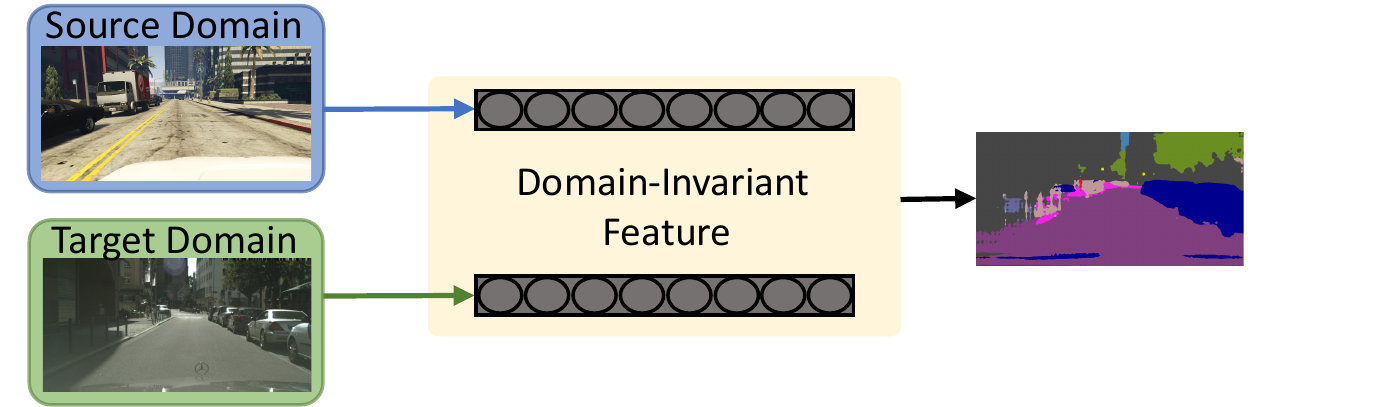} \\
        (a) Conventional domain adaptation \\
        \includegraphics[width=1\linewidth]{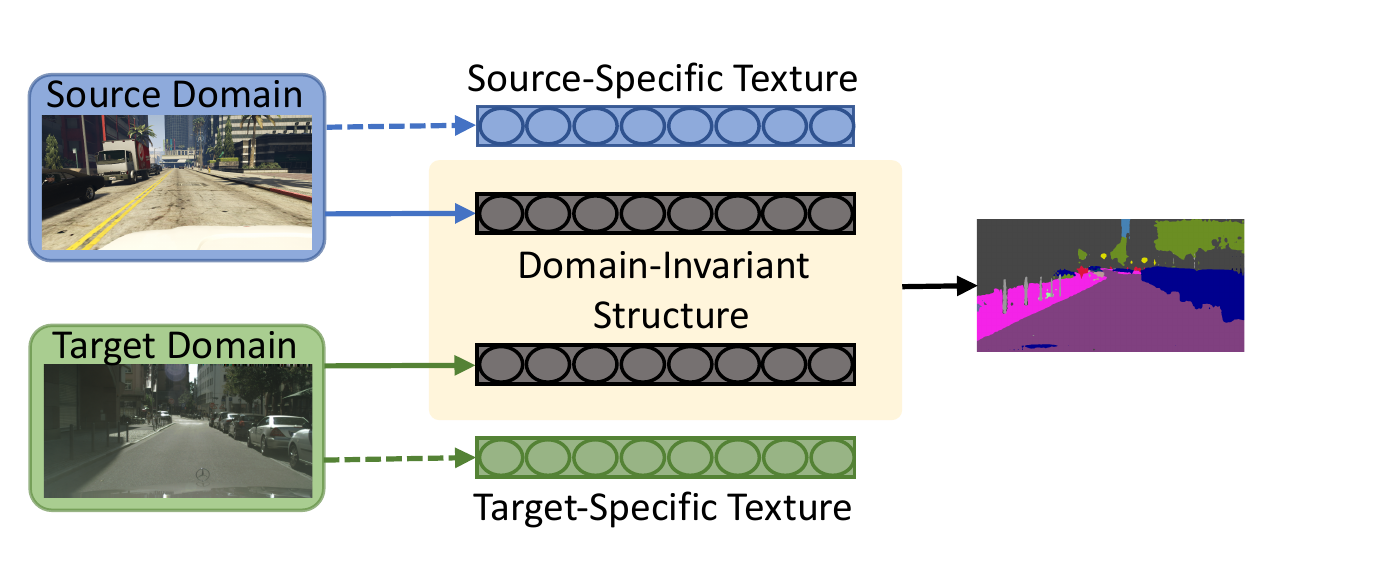} \\
        (b) The proposed method 
    \end{tabular}
    \caption{Comparison of the conventional domain adaptation for semantic segmentation and our proposed method. Instead of making the entire feature representation domain invariant, we align only the distributions of the structure component across domains.}
    \label{fig:teaser_fig}
\end{figure}

Domain adaptation is thus proposed to transfer the knowledge learned from a source domain (e.g. synthetic images) to another target domain (e.g. real images). One common approach is to learn a domain-invariant feature space across domains by matching their feature distributions, where different matching criteria have been explored, e.g. minimizing the second order statistics~\cite{sun2016deep} and domain adversarial training~\cite{ganin2014unsupervised, Hoffman_cycada2017, tzeng2017adversarial}
. There is also a recent research work~\cite{tsai2018learning} which introduces distribution alignment directly in the structural output space for the task of semantic segmentation.
However, these approaches are all driven by a strong assumption that the entire feature or output space of two domains can be well aligned (see Figure \ref{fig:teaser_fig} (a)) to yield a domain-invariant representation that is also discriminative for the tasks in question.  



In this paper, we propose a Domain Invariant Structure Extraction (DISE) framework to address unsupervised domain adaptation for semantic segmentation. We hypothesize that the \textit{high-level structure} information of an image would be the most effective for its segmentation prediction. Thus, our DISE aims to discover a \textit{domain-invariant structure feature} by learning to disentangle \textit{domain-invariant structure} information of an image from its \textit{domain-specific texture} information, as illustrated in Figure \ref{fig:teaser_fig} (b). 

Our method distinguishes from similar prior works in (1) learning an image representation comprising explicitly a domain-invariant structure component and a domain-specific texture component, (2) making only the structure component domain invariant, and (3) allowing image-to-image translation across domains which further enables label transfer, with all achieved within one single framework. Although DISE shares some parallels with domain separation networks \cite{bousmalis2016domain} and DRIT \cite{lee2018diverse}, its emphasis on the separation of structure and texture information and the ability to translate images across domains and meanwhile maintain structures clearly highlight the novelties. Extensive experiments on standardized datasets confirm its superiority over several state-of-the-art baselines.  

\begin{figure*}[h]
    \centering
	\includegraphics[width=1\textwidth]{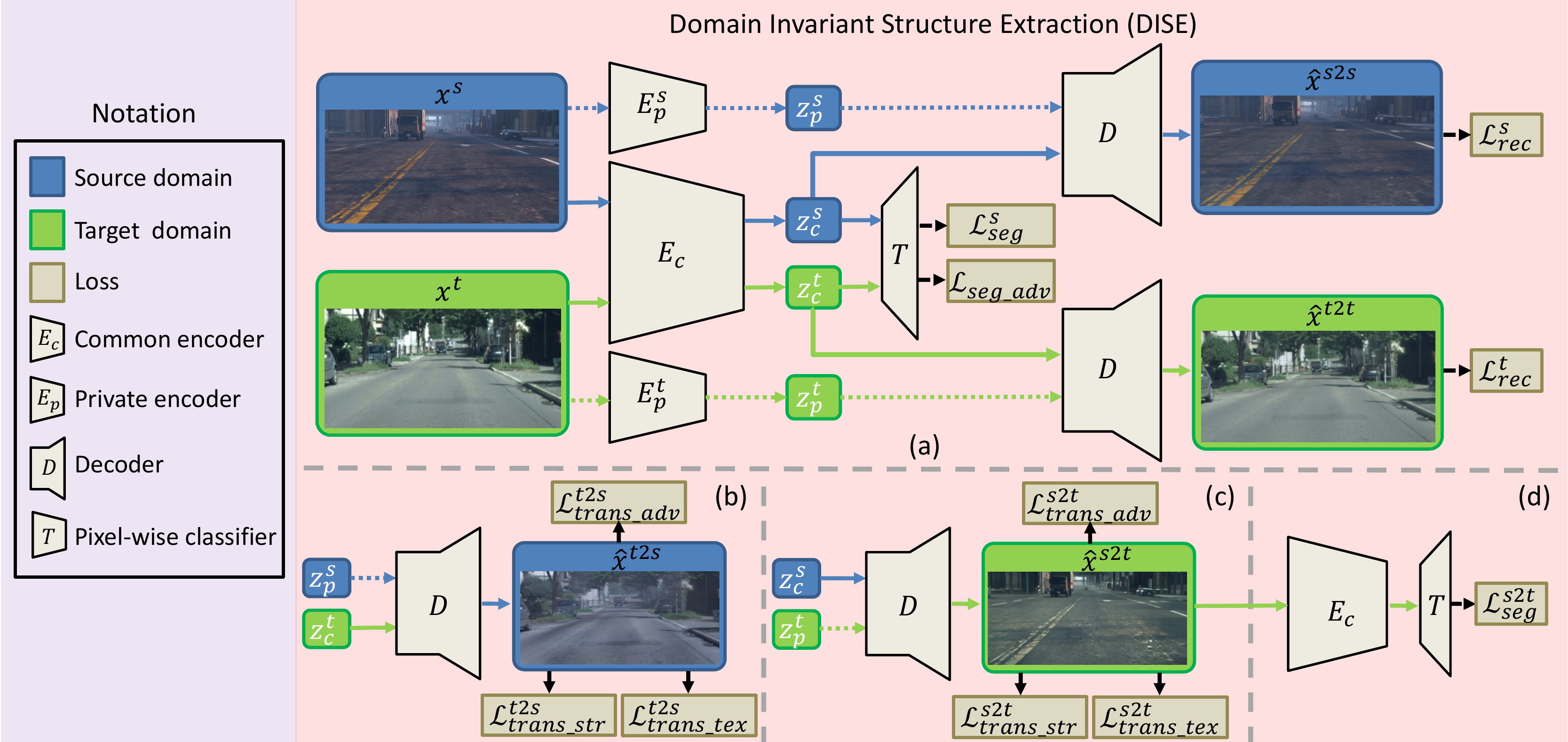}
    \caption{An overview of the proposed domain-invariant structure extraction (DISE) framework for semantic segmentation. The DISE framework is composed of a common encoder $E_c$ shared across domains, two domain-specific private encoders, $E_p^s, E_p^t$, a pixel-wise classifier $T$, and a shared decoder $D$. It encodes an image, source-domain or target-domain, into a domain-specific texture component $z_p$ and a domain-invariant structure component $z_c$, as shown in part (a). With this disentanglement, it can translate an image  $x^s$ (respectively, $x^t$) in one domain to another image $\hat{x}^{s2t}$ (respectively, $\hat{x}^{t2s}$) in the other domain by combining the structure content of $x^s$ (respectively, $x^t$) with the texture appearance of $x^t$ (respectively, $x^s$), as shown in parts (b) and (c). This further enables the transfer of ground-truth labels from the source domain to the target domain, as illustrated in part (d).} 
    \label{fig:model}
\end{figure*}

\section{Related Work}\label{sec:related}
In comparison to image classification where there exist many prior works addressing the domain adaptation problem, semantic segmentation is considered a much more challenging task to apply domain adaptation, since its output is a segmentation map full of highly structured and contextual semantic information. We review several related works here and categorize them according to the use of three widely utilized strategies: \textit{distribution alignment}, \textit{image translation}, and \textit{label transfer}. Different works may differ in their choice and conducting order of these strategies, as contrasted in Table~\ref{table:related_structure}.

Firstly, similar to the case of domain adaptation for image classification, different criteria may be applied to match distributions across domains in the feature space (e.g. \cite{hong2018conditional,sankaranarayanan2018learning,wu2018dcan,zhu2018penalizing}) or in the output space. The representative work of the latter is proposed by Tsai \etal.~\cite{tsai2018learning}, where adversarial learning is applied on segmentation maps, based on spatial contextual similarities between the source and target domains . 
However, the assumption that the whole feature or output space of the two domains can be well aligned often proves impractical, considering the substantial difference in appearance (namely, texture) between synthetic and real-world images in some applications.

Secondly, the recent advance in image-to-image translation and style transfer~\cite{huang2017arbitrary,Johnson2016Perceptual,zhu2017unpaired} has motivated the translation of source images to gain texture appearance of target images, or vice versa. On the one hand, this translation process allows segmentation models to use translated images as augmented training data~\cite{Hoffman_cycada2017, wu2018dcan}; on the other hand, the common feature space learned in the course of image translation can facilitate learning a domain-invariant segmentation model~\cite{sankaranarayanan2018learning,zhu2018penalizing}.

Finally, the image-to-image translation makes possible the transfer of labels from the source domain to the target domain, providing additional supervised signals to learn a model applicable to target-domain images~\cite{Hoffman_cycada2017, wu2018dcan}.
However, the direct image-translation may be harmful to learning, due to the risk of carrying over source specific information to the target domain. 

Our proposed DISE makes use of all three strategies but differs from these prior works in several significant ways. We hypothesize that the \textit{high-level structure} information of an image would be the most informative for its semantic segmentation. Thus, the DISE is to disentangle \textit{high-level, domain-invariant structure} information of an image from its \textit{low-level, domain-specific texture} information through a set of common and private encoders. 

\begin{table}[]
\caption{Different strategies adopted by prior works on domain adaptation for semantic segmentation. \textbf{IT}, \textbf{DA}, \textbf{LT} stand for Image Translation, Distribution Alignment, and Label Transfer, respectively. \textbf{Order} denotes the order in which these strategies are applied.}\label{table:related_structure} 
\begin{tabular}{m{0.3\linewidth}|ccc|C{0.3\linewidth}}
\hline
 Methods & \textbf{IT} & \textbf{DA} & \textbf{LT} & \textbf{Order} \\ \hline
 Sankaranarayanan \etal \cite{sankaranarayanan2018learning} &  \checkmark & \checkmark &  & IT$\rightarrow$DA\\
  Hong \etal \cite{hong2018conditional} &  & \checkmark &  & -\-- \\ 
 Wu \etal \cite{wu2018dcan} &  \checkmark &  \checkmark & \checkmark & IT$\rightarrow$DA$\rightarrow$LT \\ 
Tsai \etal \cite{tsai2018learning} &  & \checkmark &  & -\-- \\ 
 Chen \etal \cite{chen2018road} &  & \checkmark &  & -\-- \\ 
  Hoffman \etal \cite{Hoffman_cycada2017} &  \checkmark & \checkmark & \checkmark & IT$\rightarrow$LT, DA \\ 
Zhu \etal \cite{zhu2018penalizing} & \checkmark & \checkmark &  & IT$\rightarrow$DA \\ 
 Our DISE & \checkmark & \checkmark & \checkmark & DA$\rightarrow$IT$\rightarrow$LT\\ \hline
\end{tabular}
\end{table}
\section{Method}\label{sec:method}
In this paper, we propose a Domain Invariant Structure Extraction (DISE) framework to address the problem of unsupervised domain adaptation for semantic segmentation. The emphasis on explicitly regularizing the common and private encoders towards capturing structure and texture information, along with the ability to translate images from one domain to another for label transfer, underlines the novelties of our method. The following gives a formal treatment of the DISE. We begin by an overview of its framework. Next, we present in detail the loss functions used, followed by a description of implementation details. 

\subsection{Domain Invariant Structure Extraction}\label{ssec:method_DISE}
The DISE aims to learn an image representation comprising a \textit{domain-invariant structure} component and a \textit{domain-specific texture} component. The setting assumes access to $N_s$ annotated source-domain images $X^s = \{(x^{s}_{i}, y^{s}_{i})\}^{N_s}_{i=1}$, with each image $x^s_i \in \mathbb{R}^{H \times W \times 3}$ having height $H$, width $W$ and $C$-way per-pixel label of object categories $y^s_i \in \{0, 1\}^{H \times W \times C}$, and $N_t$ unannotated target-domain images $X^{t} = \{x^{t}_{i}\}^{N_t}_{i=1}$. As shown in Figure \ref{fig:model} (a), there are five sub-networks in DISE, namely, the common encoder $E_c$ shared across domains, the domain-specific private encoders $\left \{E_p^s, E_p^t\right \}$, the shared decoder $D$, and the pixel-wise classifier $T$. They are parameterized by $\theta_c$, $\theta_p^s$, $\theta_p^t$, $\theta_d$ and $\theta_t$, respectively.

Given a source-domain image $x^s$ as input, the common encoder $E_{c}$ produces $z_c^s=E_{c}(x^s; \theta_c)$ to characterize its \textit{domain-invariant, high-level structure} information while the source-specific private encoder $E_p^s$ generates $z_p^s=E_{p}^{s}(x^s;\theta_p^s)$ for capturing its remnant aspects that are largely related to \textit{domain-specific, low-level texture} information. These two components $\left \{z_c^s, z_p^s\right \}$ are complementary to each other; when combined together, they allow the decoder $D$ to minimize a reconstruction loss $\mathcal{L}_{rec}^s$ between the input $x^s$ and its reconstruction $\hat{x}^{s2s} = D(z_c^s, z_p^s; \theta_d)$. Likewise, a target-domain image $x^t$ can be encoded and decoded similarly to minimize $\mathcal{L}_{rec}^t$, yielding $z_c^t=E_{c}(x^t;\theta_c),z_p^t=E_{p}^{t}(x^t;\theta_p^t)$ and $x^t \approx \hat{x}^{t2t} = D(z_c^t, z_p^t; \theta_d)$, where
the private encoder $E_{p}^{t}$, like its counterpart $E_{p}^{s}$, extracts the target-specific texture information. It is the structure components $z_c^s,z_c^t$ that will be used by classifier $T$ to predict segmentation maps, $\hat{y}^s = T(z_c^s,\theta_t),\hat{y}^t = 
T(z_c^t,\theta_t)$ in source and target domains accordingly. 

The disentanglement between structure and texture information is realized by the regularization coming from image translation with domain adversarial training \cite{liu2017unsupervised} and perceptual loss minimization \cite{Johnson2016Perceptual}. 
As illustrated in Figure \ref{fig:model} (b) and (c), we consider any pair of source- and target-domain images with their respective representations $x^s=\{z_c^s, z_p^s\}$ and $x^t=\{z_c^t, z_p^t\}$. We first interchange their domain-specific components, and then decode them into two unseen, translated images $\hat{x}^{s2t} = D(z_c^s, z_p^t; \theta_d)$ and $\hat{x}^{t2s} = D(z_c^t, z_p^s; \theta_d)$. If the common and private encoders behave as we expect them to capture the structure and texture information, respectively, the translated image $\hat{x}^{s2t}$ (respectively, $\hat{x}^{t2s}$) should hold the high-level structure the same as $x^s$ (respectively, $x^t$) while exhibiting similar low-level texture appearance to $x^t$ (respectively, $x^s$). To this end, we train our networks by imposing domain adversarial losses $\mathcal{L}_{trans\_adv}^{t2s},\mathcal{L}_{trans\_adv}^{s2t}$ \cite{liu2017unsupervised} and perceptual losses $\mathcal{L}_{trans\_str}^{t2s},\mathcal{L}_{trans\_tex}^{t2s},\mathcal{L}_{trans\_str}^{s2t},\mathcal{L}_{trans\_tex}^{s2t}$ \cite{Johnson2016Perceptual} at the output of the decoder $D$ in order to ensure the domain and perceptual similarities between these translated images and their counterparts in the source or target domains. This image translation functionality of DISE further allows the transfer of ground-truth labels from the source domain to the target domain. More specifically, since the target-domain-like images $\hat{x}^{s2t}$ share the same structure component as $x^s$, we consider the ground-truth labels $y^s$ of $x^s$ to be the pseudo labels for $\hat{x}^{s2t}$ on grounds of our hypothesis that the segmentation prediction for an image depends solely on its structure information.

Finally, we make the structure components $z_c^s,z_c^t$ invariant to the domain from which they are extracted by minimizing another domain adversarial loss $\mathcal{L}_{seg\_adv}$ at the output of the classifier $T$, as well as the negative log-likelihood functions of the ground-truth labels $y^s$ with respect to $x^s$ and $\hat{x}^{s2t}$, i.e. $\mathcal{L}_{seg}^s$ and $\mathcal{L}_{seg}^{s2t}$ (see Figure \ref{fig:model} (d)).

\subsection{Learning}\label{ssec:Learning}
The training of the proposed DISE is to minimize a weighted combination of the aforementioned loss functions with respect to the parameters $\{\theta_c, \theta_p^s, \theta_p^t, \theta_d\, \theta_t\}$ of the five sub-networks:
\begin{equation}
    \begin{aligned}
	    \mathcal{L}  = 
	    & \lambda_{seg}^s~\mathcal{L}_{seg}^s+\lambda_{seg\_adv}~\mathcal{L}_{seg\_adv}+\lambda_{rec}~\mathcal{L}_{rec}
	    \\& +\lambda_{trans\_str}~\mathcal{L}_{trans\_str} +\lambda_{trans\_tex}~\mathcal{L}_{trans\_tex}
	    \\& +\lambda_{trans\_adv}~\mathcal{L}_{trans\_adv}+\lambda_{seg}^{s2t}~\mathcal{L}_{seg}^{s2t},
    \end{aligned}
\end{equation}
where the combination weights $\lambda$'s are chosen empirically to strike a balance among the model capacity, reconstruction/translation quality, and prediction accuracy. In the following, we elaborate on each of these loss functions.
\vspace{0.5em}\\  
\textbf{Segmentation Loss.} The segmentation loss $\mathcal{L}_{seg}^s(\theta_c, \theta_t)$ given by the typical cross-entropy based on the source-domain ground truths $y^s$ is to train supervisedly the common encoder $E_c$ and the classifier $T$ in order to predict segmentation maps $\hat{y}^s$ for source-domain images $x^s$. 
\vspace{0.5em}\\  
\textbf{Output Space Adversarial Loss.} Inspired by Tsai \etal \cite{tsai2018learning}, we introduce an adversarial loss $\mathcal{L}_{seg\_adv}(\theta_c, \theta_t)$ at the output of the classifier $T$, in the hopes of making the common encoder $E_c$ and the classifier $T$ generalize well on target-domain images. Specifically, we first train a discriminator $D_{adv}^{seg}$ to distinguish between the source prediction $\hat{y}^s$ and the target prediction $\hat{y}^t$ at the patch level \cite{isola2017image} by minimizing a supervised domain loss (i.e. $D_{adv}^{seg}$ should ideally output 1 for each patch in the source prediction $\hat{y}^s$ and 0 for that in the target prediction $\hat{y}^t$).
We then update the common encoder $E_c$ and the classifier $T$ to fool the discriminator $D_{adv}^{seg}$ by inverting its output for $\hat{y}^t$ from 0 to 1, that is, by minimizing 
\begin{equation}
    \begin{aligned}
        \mathcal{L}_{seg\_adv}(\theta_c, \theta_t) = -\frac{1}{H'W'} \sum_{h', w'}\log(D_{adv}^{seg}(\hat{y}^t)_{h',w'}),
    \end{aligned}
    \label{Eq:seg_adv}
\end{equation}
where $h',w'$ are patch coordinates and $H'=H/16, W'=W/16$ with the factor 16 accounting for the downsampling in the discriminator $D_{adv}^{seg}$.
\vspace{0.5em} \\ 
\textbf{Reconstruction Loss.} The reconstruction loss $\mathcal{L}_{rec}(\theta_c, \theta_p^s,\theta_p^t,\theta_d)$ is to ensure that the two domain-invariant and domain-specific components $z_c, z_p$ of an image representation together form a nearly complete summary of the image. 
To encourage the reconstruction to be perceptually similar to the input image, we follow the notion of perceptual loss \cite{Johnson2016Perceptual} to define our quality metric $\mathcal{L}_{perc}(x,y;w)$ as a weighted sum of L1 differences between feature representations extracted from a pre-trained VGG network \cite{Simonyan14c}. In symbols, we have 
\begin{equation}
    \begin{aligned}
    	\mathcal{L}_{perc}(x, y; w) = \sum_{l\in L}\frac{w^{(l)}}{N^{(l)}}\left\|\psi^{(l)}(x)-\psi^{(l)}(y)\right\|_1,
    \end{aligned}
    \label{Eq:perc}
\end{equation}
where $\psi^{(l)}(x)$ (respectively, $\psi^{(l)}(y)$) is the activations of the $l$-th layer of the pre-trained VGG network for input $x$ (respectively, $y$), $N^{(l)}$ is the number of activations in layer $l$, $w^{(l)}$ gives a separate weighting to the loss in layer $l$, and $L$ refers to $\{ \texttt{relu1\_1}, \texttt{relu2\_1}, \texttt{relu3\_1}, \texttt{relu4\_1}, \texttt{relu5\_1}\}$ of the VGG network.
As pointed out in~\cite{Johnson2016Perceptual}, the higher layers of VGG network tend to represent the high-level structure content of an image while the lower layers generally describe its low-level texture appearance. Equation \ref{Eq:perc} is then used to regularize the reconstruction of both source- and target-domain images by minimizing the sum of their respective perceptual losses:
\begin{equation}
    \begin{aligned}
        & \mathcal{L}_{rec}(\theta_c, \theta_p^s,\theta_p^t,\theta_d)   \\ & = \mathcal{L}_{rec}^s + \mathcal{L}_{rec}^t
        \\ & =  \mathcal{L}_{perc}(\hat{x}^{s2s}, x^s;w_{rec}) + \mathcal{L}_{perc}(\hat{x}^{t2t}, x^t;w_{rec}),
    \end{aligned}
    \label{Eq:recon}
\end{equation} 
where the weighting $w_{rec}$ is set to weight more on higher layers.
\vspace{0.5em} \\
\textbf{Translation Structure Loss.} As motivated previously in Section~\ref{ssec:method_DISE}, an image produced by translation across domains should keep its structure unchanged.
The translation structure loss $\mathcal{L}_{trans\_str}(\theta_c,\theta_p^s,\theta_p^t,\theta_d)$ as defined in Equation \ref{Eq:trans_str} measures the differences in high-level structure between the translated image $\hat{x}^{s2t}$ and the image $x^s$ from which the structure component of $\hat{x}^{s2t}$ is derived, and likewise, between $\hat{x}^{t2s}$ and $x^t$. This is achieved by choosing for the perceptual metric a weighting $w_{str}$ that again stresses on the feature reconstruction losses in higher layers of the pre-trained VGG network. Our goal is to penalize the translated images which differ significantly in structure from the images with which they share the same structure component $z_c$, thereby getting $z_c$ to encode explicitly the structure aspect of an image. 
\begin{equation}
    \begin{aligned}
        & \mathcal{L}_{trans\_str}(\theta_c,\theta_p^s,\theta_p^t,\theta_d) 
        \\ & = \mathcal{L}_{trans\_str}^{s2t}+\mathcal{L}_{trans\_str}^{t2s}
        \\ & = \mathcal{L}_{perc}(\hat{x}^{s2t}, x^s; w_{str})  + \mathcal{L}_{perc}(\hat{x}^{t2s}, x^t; w_{str})
    \end{aligned}
    \label{Eq:trans_str}
\end{equation}
\vspace{0.5em}\\ 
\textbf{Translation Texture Loss.} The translation texture loss $\mathcal{L}_{trans\_tex}(\theta_c,\theta_p^s,\theta_p^t,\theta_d)$ further requires that the translated image $\hat{x}^{s2t}$  (respectively, $\hat{x}^{t2s}$) should resemble closely in texture the image $x^t$ (respectively, $x^s$), since they share the same texture component $z_p$. In doing so, $z_p$ has to encode explicitly the texture aspect of an image. Inspired by the work of AdaIN \cite{huang2017arbitrary}, we propose a weighted metric $\mathcal{L}_{tex}(x, y; w)$ to measure channel-wisely the difference in the mean value of their activations extracted from a pre-trained VGG network: 
\begin{equation}
        \begin{aligned}
        	& \mathcal{L}_{tex}(x, y; w) 
            \\ & = \sum_{l \in L} \frac{w^{(l)}}{C^{(l)}} 
        	\sum_c \left\|\mu_c(\psi^{(l)}(x))-\mu_c(\psi^{(l)}(y))\right\|_1,
        \end{aligned}
        \label{Eq:tex_loss}
\end{equation}
where $C^{(l)}$ is the number of channels in layer $l$ of the VGG network, $w^{(l)}$ specifies the weighting given to layer $l$, and $\mu_c(\cdot)$ returns the mean activation of channel $c$. Like the translation structure loss, the translation texture loss also involves the two types of translation:
\begin{equation}
    \begin{aligned}
    & \mathcal{L}_{trans\_tex}(\theta_c,\theta_p^s,\theta_p^t,\theta_d) 
    \\ & = \mathcal{L}_{trans\_tex}^{s2t}+\mathcal{L}_{trans\_tex}^{t2s}
    \\ & = \mathcal{L}_{tex}(\hat{x}^{s2t}, x^t; w_{tex})  + \mathcal{L}_{tex}(\hat{x}^{t2s}, x^s; w_{tex}),
    \end{aligned}
\end{equation} 
where the weighting $w_{tex}$ of the perceptual metric is now chosen to emphasize more on early layers.
\vspace{0.5em}\\  
\textbf{Translation Adversarial Loss.} In addition to the aforementioned perceptual losses, we also employ adversarial losses $\mathcal{L}_{trans\_adv}(\theta_c,\theta_p^s,\theta_p^t,\theta_d)$ to adapt the translated images $\hat{x}^{s2t}$ and $\hat{x}^{t2s}$ to appear as if they were images out of the target and source domains, respectively. To this end, we adopt LSGAN \cite{mao2017least} and Patch Discriminator \cite{isola2017image}.
\vspace{0.5em}\\ 
\textbf{Label Transfer Loss.} The label transfer loss $\mathcal{L}_{seg}^{s2t}(\theta_c,\theta_t)$ is given by a typical cross-entropy loss that trains supervisedly the common encoder $E_c$ and the classifer $T$ on translated images $\hat{x}^{s2t}$ with pseudo labels $y^s$. 
\subsection{Implementation}\label{ssec:method_network}
\paragraph{Networks.} For experiments, we use a base model, referring collectively to the common encoder $E_c$ and the pixel-wise classifier $T$, similar to the segmentation network in \cite{tsai2018learning}, which is built on DeepLab-v2 \cite{chen2018deeplab} with ResNet-101 \cite{he2016deep}. We obtain initial weights by pre-training on PASCAL VOC \cite{everingham2015pascal} dataset, and at training time, reuse the pre-trained batchnorm layer. The common encoder $E_c$ outputs the feature maps of the last residual layer ($layer4$) as $z_c$. For the private encoders $E_p^s,E_p^t$, we adopt a convolutional neural network containing 4 convolution blocks, followed by one global pooling layer and one fully-connected layer. The output of the private encoder $E_p^s$ (respectively, $E_p^t$) is an 8-dimensional representation $z_p^s$ (respectively, $z_p^t$). For the shared decoder $D$, we use three residual blocks and three deconvolution layers. The input to the decoder is a concatenation of the private code $z_p$, the feature maps $z_c$, and a flag indicating the domain of the private code. 

\paragraph{Training Details.} We implement DISE with Pytorch on a single Tesla V100 with 16 GB memory. The full training takes 88 GPU hours. Due to limited memory, at training time, we resize input images to 512$\times$1024 and perform random cropping with a crop size of 256$\times$512. However, at test time, the input images are of size 512$\times$1024. For fair comparison, we follow Tsai \etal~\cite{tsai2018learning} and resize the output predictions from 512$\times$1024 to 1024$\times$2048 at evaluation time. We train our model for 250,000 iterations with a batch size of 2. We use the SGD solver with an initial learning rate of $2.5\times10^{-4}$ for the common encoder $E_c$ and the classifier $T$; the Adam solver with an initial learning rate of $1.0\times10^{-3}$ for the decoder $D$; and the Adam solver with an initial learning rate of $1.0\times10^{-4}$ for the others. All the learning rates decrease according to the polynomial decay policy. The momentum is set to 0.9 and 0.99.
\section{Experimental Results}\label{sec:exp}
In this section, we perform experiments on typical datasets for semantic segmentation. We compare the performance of our proposed method with several state-of-the-art baselines and conduct an ablation study to understand the effect of various combinations of loss functions on segmentation performance. The code and pre-trained models are available online\footnote{\url{https://github.com/a514514772/DISE-Domain-Invariant-Structure-Extraction}}.  



\subsection{Datasets}\label{ssec:exp_dataset}
For experiments, we follow the common protocol adopted by most prior works; that is, taking synthetic dataset GTA5~\cite{richter2016playing} or SYNTHIA~\cite{ros2016synthia} with ground-truth annotations as the source domain, and Cityscapes dataset~\cite{cordts2016cityscapes} as the target domain where no annotation is available during training. At test time, the evaluation is conducted on the validation set of Cityscapes. The details of these datasets are described as follows. 
\vspace{0.5em} \\
\textbf{Cityscapes} \cite{cordts2016cityscapes} is a real-world dataset composed of street-view images captured in 50 different cities. Its data split includes 2975 training images and 500 validation images, with each having a spatial resolution of 2048 $\times$ 1024 and 19 semantic labels at the pixel level. Note again that no ground-truth label is used in model training. 
\vspace{0.5em} \\
\textbf{GTA5} \cite{richter2016playing} is a synthetic dataset containing 24996 images of size 1914 $\times$ 1052. These images are collected from computer game Grand Theft Auto V (GTAV) and come with pixel-level semantic labels that are fully compatible with Cityscapes \cite{cordts2016cityscapes}. 
\vspace{0.5em} \\
\textbf{SYNTHIA} is another synthetic dataset composed of 9400 annotated synthetic images with the resolution 1280 $\times$ 960. Like GTA5, it has semantically compatible annotations with Cityscapes \cite{cordts2016cityscapes}. Following the prior works \cite{hong2018conditional, sankaranarayanan2018learning, tsai2018learning, wu2018dcan}, we use the SYNTHIA-RAND-CITYSCAPE subset \cite{ros2016synthia}. 

\subsection{Performance Comparison}\label{ssec:exp_cmp_sota}
We compare the performance of our method against several baselines, including the models of~\cite{chen2018road, hong2018conditional, saleh2018effective, sankaranarayanan2018learning, tsai2018learning, wu2018dcan}. 
Of these, the works \cite{chen2018road, hong2018conditional, tsai2018learning} are representative of the conventional adaptation that matches distributions of feature or output spaces across domains based on adversarial training; the works~\cite{sankaranarayanan2018learning, wu2018dcan} are typical of those that map source-domain images to the target domain at the pixel level by image translation or style transfer; and Saleh \etal~\cite{saleh2018effective} stands out from the others by object detection-based method for foreground instances. More details of these works can be found in Section \ref{sec:related}.
\vspace{0.5em} \\
\textbf{GTA5 to Cityscapes.} Table \ref{table:performance} shows that as compared to the baselines, our method achieves the state-of-the-art performance of 45.4 in mean intersection-over-union (mIoU). A breakdown analysis further reveals that it outperforms most of the baselines by a large margin in predicting "Road", "Sidewalk, "Wall", "Fence", "Building", and "Sky" classes. These are classes that often appear concurrently in an image and tend to be spatially connected. Moreover, some of them, e.g. "Road" and "Sidewalk", exhibit highly similar texture appearance. We thus attribute the good performance of our scheme to its ability to filter out the domain-specific texture information in forming a domain-invariant structure representation for semantic segmentation. 

In Figure \ref{fig:exp_result}, we show qualitative results comparing our method against "Source Only" (i.e. no adaptation) and "Conventional Adaptation" (i.e. without disentanglement of structure and texture). For the latter, we present results of \cite{tsai2018learning}. It is clear that the segmentation predictions made by our method look most similar to the ground truths. On closer examination, we see that our model can better discern the difference between "Sidewalk" and "Road" as compared to the baselines. It also does a good job at identifying rare classes such as "Pole" and "Traffic Sign". These observations suggest that our structure-based representations are indeed more discriminative than other representations that may have encoded both structure and texture information as with the "Conventional Adaptation".    
\vspace{0.5em} \\
\textbf{SYNTHIA to Cityscapes.} 
We also evaluate all models on the more challenging SYNTHIA dataset. Specifically, we follow \cite{tsai2018learning} to compare results based on semantic predictions for only 16 classes. Table \ref{table:performance_synthia} presents quantitative results in terms of per-class IoU and mIoU. It is seen that most of the aforementioned discussions made with GTA5 dataset can be carried over to SYNTHIA. Although the prior work \cite{hong2018conditional} performs closely to our model in terms of mIoU, the superiority of our method in classes like "Road", "Sidewalk", "Building", "Sky" still remains.    

\begin{table*}[bt]
\centering\footnotesize
\setlength\tabcolsep{0.25em}
\caption{Comparison results on Cityscapes when adapted from GTA5 in terms of per-class IoU and mIoU over 19 classes.}\label{table:performance}
\begin{tabular}{@{}m{2.1cm}|m{2cm}|*{19}{c}|c@{}}
\hline
Methods                  & Base Model  & \rot{Road} & \rot{Sidewalk} & \rot{Building} & \rot{Wall} & \rot{Fence} & \rot{Pole} & \rot{Traffic Light } & \rot{Traffic Sign} & \rot{Vegetation} & \rot{Terrain} & \rot{Sky}  & \rot{Person} & \rot{Rider} & \rot{Car}  & \rot{Truck} & \rot{Bus}  & \rot{Train} & \rot{Motorbike} & \rot{Bicycle} & \rot{mIoU} \\ \hline
Sankaranarayanan \etal. \cite{sankaranarayanan2018learning}& FCN8s \cite{long2015fully} & 88.0 & 30.5 & 78.6 & 25.2 & 23.5 & 16.7 & 23.5 & 11.3 & 78.7 & 27.2 & 71.9 & 51.3 & 19.5 & 80.4 & 19.8 & 18.3 & 0.9 & 20.8 & 18.4 & 37.1 \\
Wu \etal. \cite{wu2018dcan}& FCN8s \cite{long2015fully} & 88.5 & 37.4 & 79.3 & 24.8 & 16.5 & 21.3 & 26.3 & 17.4 & 80.8 & 30.9 & 77.6 & 50.2 & 19.2 & 77.7 & 21.6 & 27.1 & 2.7 & 14.3 & 18.1 & 38.5 \\
Hong \etal. \cite{hong2018conditional}& FCN8s \cite{long2015fully} & 89.2 & \textbf{49.0} & 70.7 & 13.5 & 10.9 & \textbf{38.5} & 29.4 & 33.7 & 77.9 & 37.6 & 65.8 & \textbf{75.1} & \textbf{32.4} & 77.8 & \textbf{39.2} & 45.2 & 0.0 & 25.5 & \textbf{35.4} & 44.5 \\
Chen \etal. \cite{chen2018road}& PSPNet \cite{zhao2017pyramid}& 76.3 & 36.1 & 69.6 & 28.6 & 22.4 & 28.6 & 29.3 & 14.8 & 82.3 & 35.3 & 72.9 & 54.4 & 17.8 & 78.9 & 27.7 & 30.3 & 4.0 & 24.9 & 12.6 & 39.4 \\
Wu \etal. \cite{wu2018dcan}& PSPNet \cite{zhao2017pyramid}& 85.0 & 30.8 & 81.3 & 25.8 & 21.2 & 22.2 & 25.4 & 26.6 & \textbf{83.4} & 36.7 & 76.2 & 58.9 & 24.9 & 80.7 & 29.5 & \textbf{42.9} & 2.5 & 26.9 & 11.6 & 41.7 \\
Chen \etal. \cite{chen2018road}& Deeplab v2 \cite{chen2018deeplab}& 85.4 & 31.2 & 78.6 & 27.9 & 22.2 & 21.9 & 23.7 & 11.4 & 80.7 & 29.3 & 68.9 & 48.5 & 14.1 & 78.0 & 19.1 & 23.8 & 9.4 & 8.3 & 0.0 & 35.9 \\
Tsai \etal. \cite{tsai2018learning}& Deeplab v2 \cite{chen2018deeplab}& 86.5 & 36.0 & 79.9 & 23.4 & 23.3 & 23.9 & \textbf{35.2} & 14.8 & \textbf{83.4} & 33.3 & 75.6 & 58.5 & 27.6 & 73.7 & 32.5 & 35.4 & 3.9 & \textbf{30.1} & 28.1 & 42.4 \\
Saleh \etal. \cite{saleh2018effective}& Deeplab v2 \cite{chen2018deeplab}& 79.8 & 29.3 & 77.8 & 24.2 & 21.6 & 6.9 & 23.5 & \textbf{44.2} & 80.5 & \textbf{38.0} & 76.2 & 52.7 & 22.2 & 83.0 & 32.3 & 41.3 & \textbf{27.0} & 19.3 & 27.7 & 42.5 \\ \hline
Ours & Deeplab v2 \cite{chen2018deeplab}& \textbf{91.5} & 47.5 & \textbf{82.5} & \textbf{31.3} & \textbf{25.6} & 33.0 & 33.7 & 25.8 & 82.7 & 28.8 & \textbf{82.7} & 62.4 & 30.8 & \textbf{85.2} & 27.7 & 34.5 & 6.4 & 25.2 & 24.4 & \textbf{45.4} \\ \hline
\end{tabular}
\end{table*}

\begin{table*}[]
\centering\footnotesize
\setlength\tabcolsep{0.45em}
\caption{Comparison results on Cityscapes when adapted from SYNTHIA in terms of per-class IoU and mIoU over 16 classes.}\label{table:performance_synthia}
\begin{tabular}{@{}m{2.1cm}|m{1.9cm}|*{16}{c}|c@{}}
\hline
Methods & Base Model & \rot{Road} & \rot{Sidewalk} & \rot{Building} & \rot{Wall} & \rot{Fence} & \rot{Pole} & \rot{Traffic Light } & \rot{Traffic Sign} & \rot{Vegetation} & \rot{Sky} & \rot{Person} & \rot{Rider} & \rot{Car} & \rot{Bus} & \rot{Motorbike} & \rot{Bicycle} & \rot{mIoU} \\ \hline
Sankaranarayanan \etal. \cite{sankaranarayanan2018learning} & FCN8s \cite{long2015fully} & 80.1 & 29.1 & \textbf{77.5} & 2.8 & 0.4 & 26.8 & 11.1 & 18.0 & 78.1 & 76.7 & 48.2 & 15.2 & 70.5 & 17.4 & 8.7 & 16.7 & 36.1 \\
Wu \etal. \cite{wu2018dcan} & FCN8s \cite{long2015fully} & 81.5 & 33.4 & 72.4 & 7.9 & 0.2 & 20.0 & 8.6 & 10.5 & 71.0 & 68.7 & 51.5 & 18.7 & 75.3 & 22.7 & 12.8 & 28.1 & 36.5 \\
Hong \etal. \cite{hong2018conditional} & FCN8s \cite{long2015fully} & 85.0 & 25.8 & 73.5 & 3.4 & \textbf{3.0} & \textbf{31.5} & \textbf{19.5} & \textbf{21.3} & 67.4 & 69.4 & \textbf{68.5} & \textbf{25.0} & 76.5 & \textbf{41.6}& 17.9 & 29.5 & 41.2 \\
Wu \etal. \cite{wu2018dcan} & PSPNet \cite{zhao2017pyramid} & 82.8 & 36.4 & 75.7 & 5.1 & 0.1 & 25.8 & 8.04 & 18.7 & 74.7 & 76.9 & 51.1 & 15.9 & 77.7 & 24.8 & 4.1 & \textbf{37.3} & 38.4 \\
Chen \etal. \cite{chen2018road}& Deeplab v2 \cite{chen2018deeplab} & 77.7 & 30.0 & \textbf{77.5} & \textbf{9.6} & 0.3 & 25.8 & 10.3 & 15.6 & 77.6 & 79.8 & 44.5 & 16.6 & 67.8 & 14.5 & 7.0 & 23.8 & 36.2 \\
Tsai \etal. \cite{tsai2018learning} & Deeplab v2 \cite{chen2018deeplab} & 84.3 & 42.7 & \textbf{77.5} & 9.3 & 0.2 & 22.9 & 4.7 & 7.0 & 77.9 & \textbf{82.5} & 54.3 & 21.0 & 72.3 & 32.2 & \textbf{18.9} & 32.3 & 40.0 \\ \hline
Ours & Deeplab v2 \cite{chen2018deeplab} & \textbf{91.7} & \textbf{53.5} & 77.1 & 2.5 & 0.2 & 27.1 & 6.2 & 7.6 & \textbf{78.4} & 81.2 & 55.8 & 19.2 & \textbf{82.3} & 30.3 & 17.1 & 34.3 & \textbf{41.5} \\
\hline
\end{tabular}
\end{table*}

\begin{figure*}
    \centering
    \setlength\tabcolsep{0.25em}
    \begin{tabular}{@{}ccccc@{}}
         \includegraphics[width=0.18\linewidth]{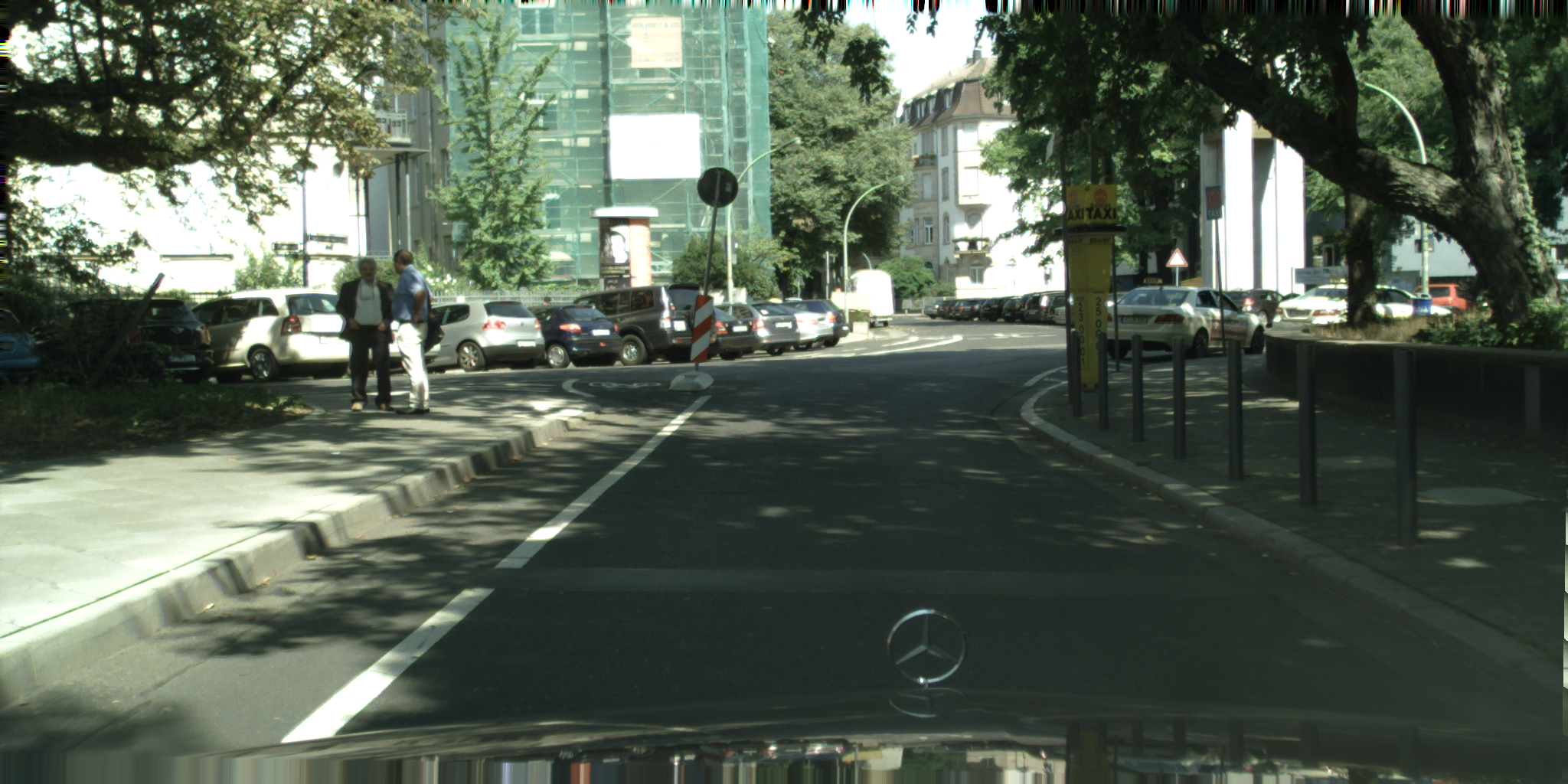} & \includegraphics[width=0.18\linewidth]{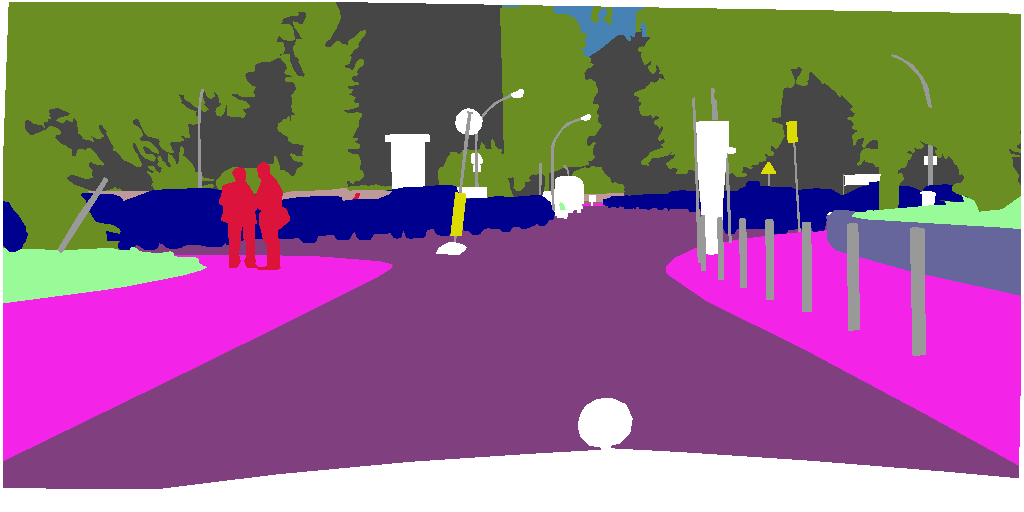} & \includegraphics[width=0.18\linewidth]{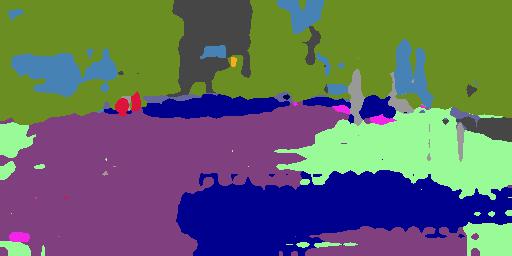} &
         \includegraphics[width=0.18\linewidth]{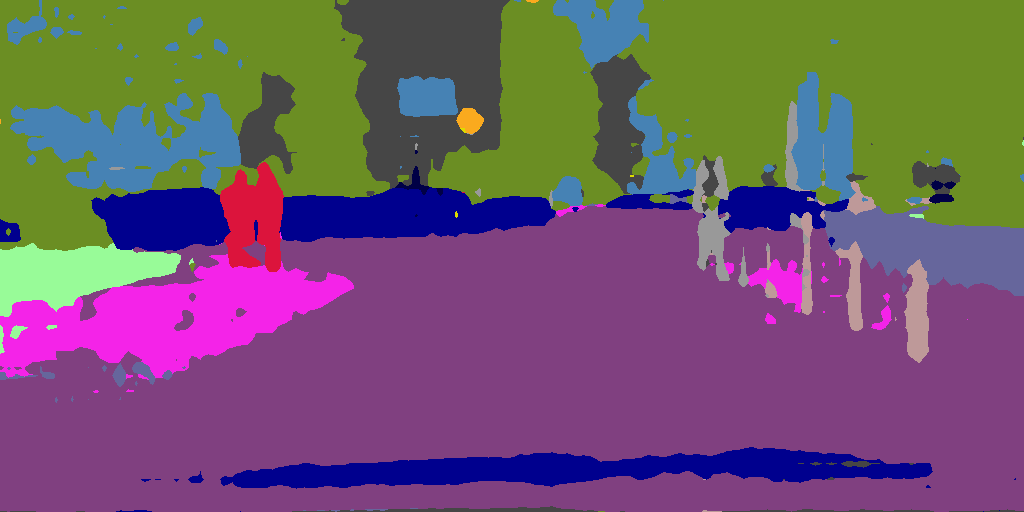} & \includegraphics[width=0.18\linewidth]{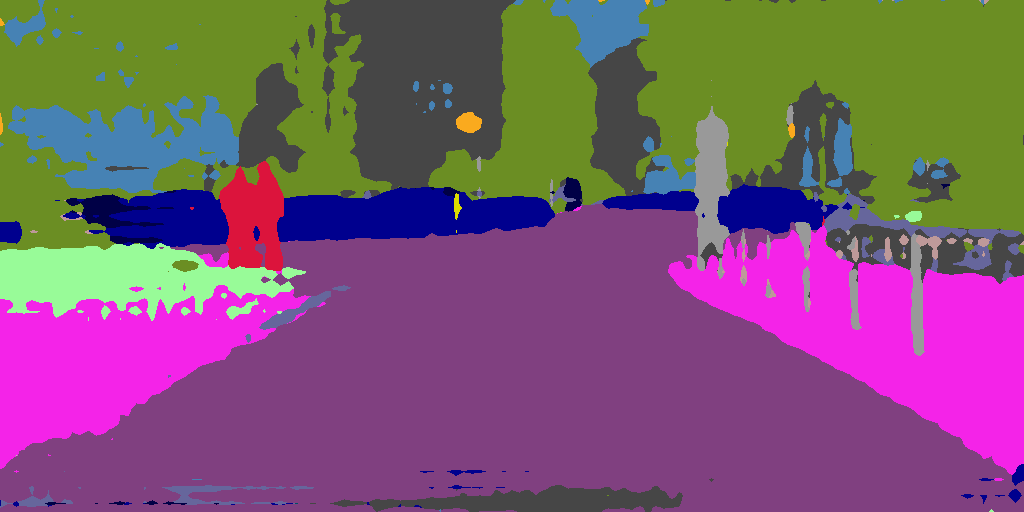} \\
         \includegraphics[width=0.18\linewidth]{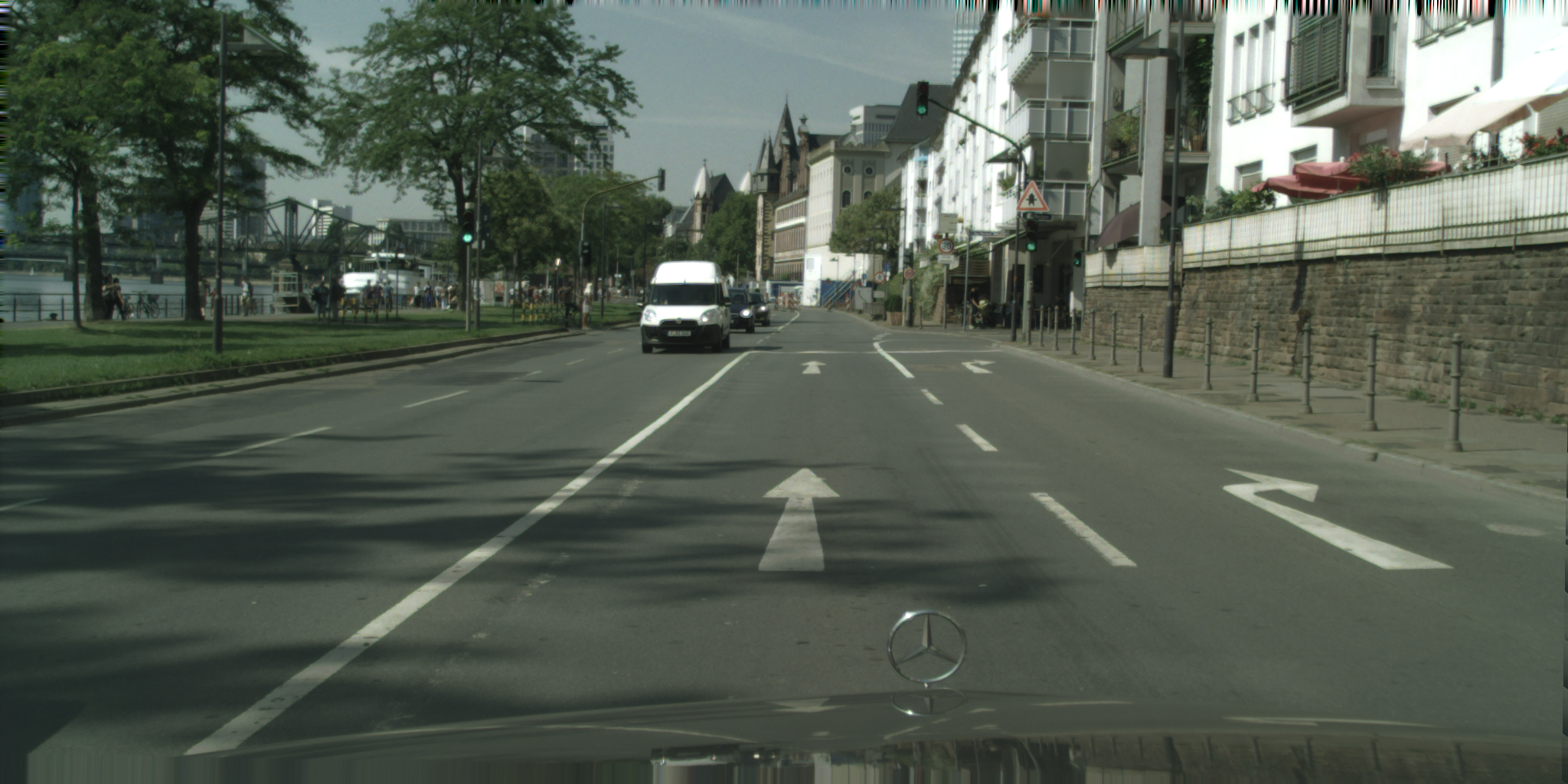} & \includegraphics[width=0.18\linewidth]{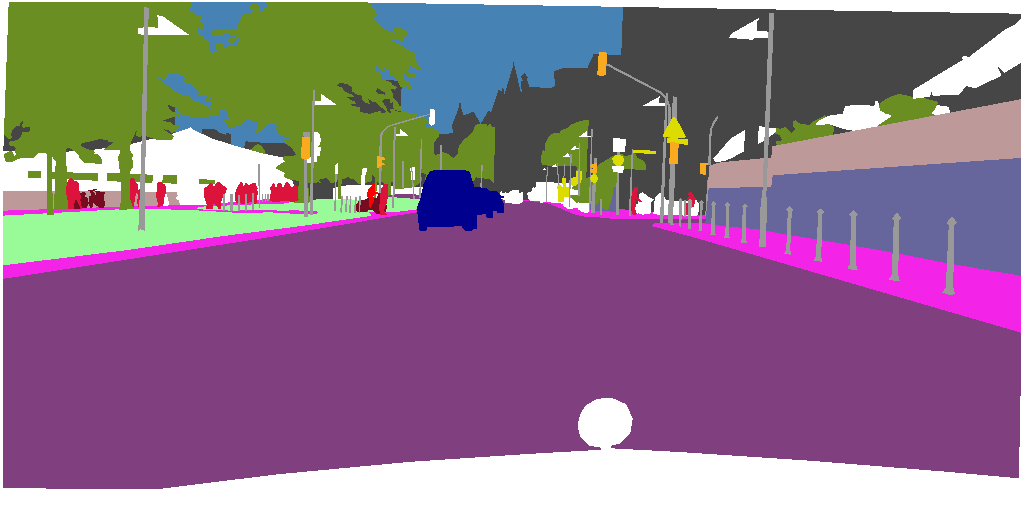} & \includegraphics[width=0.18\linewidth]{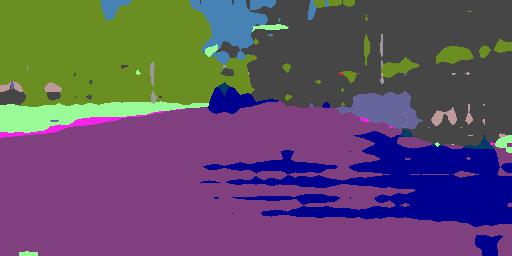} &
         \includegraphics[width=0.18\linewidth]{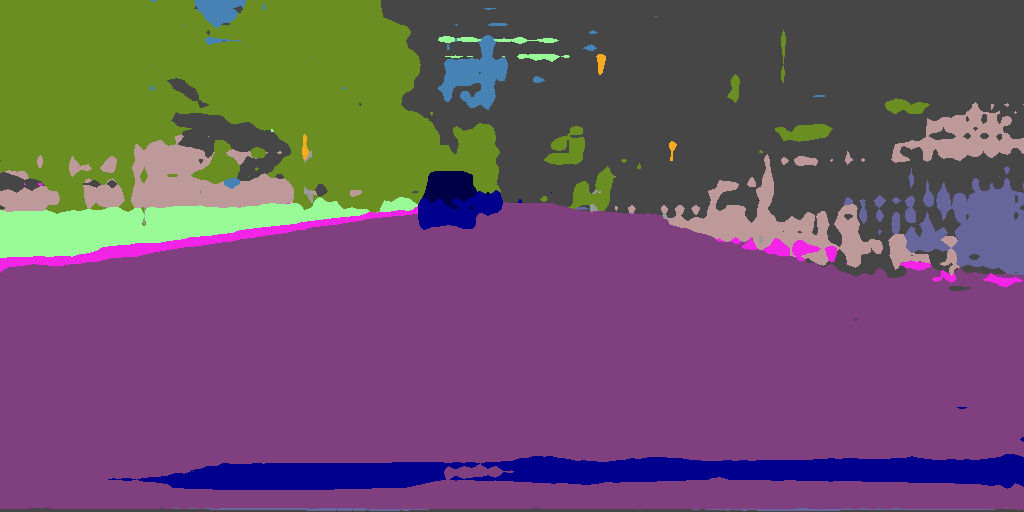} & \includegraphics[width=0.18\linewidth]{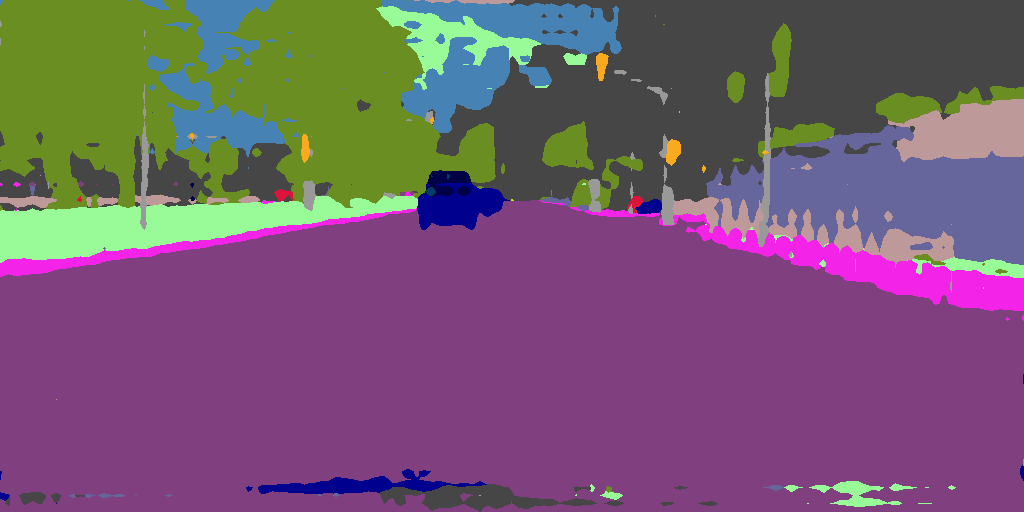} \\
         \includegraphics[width=0.18\linewidth]{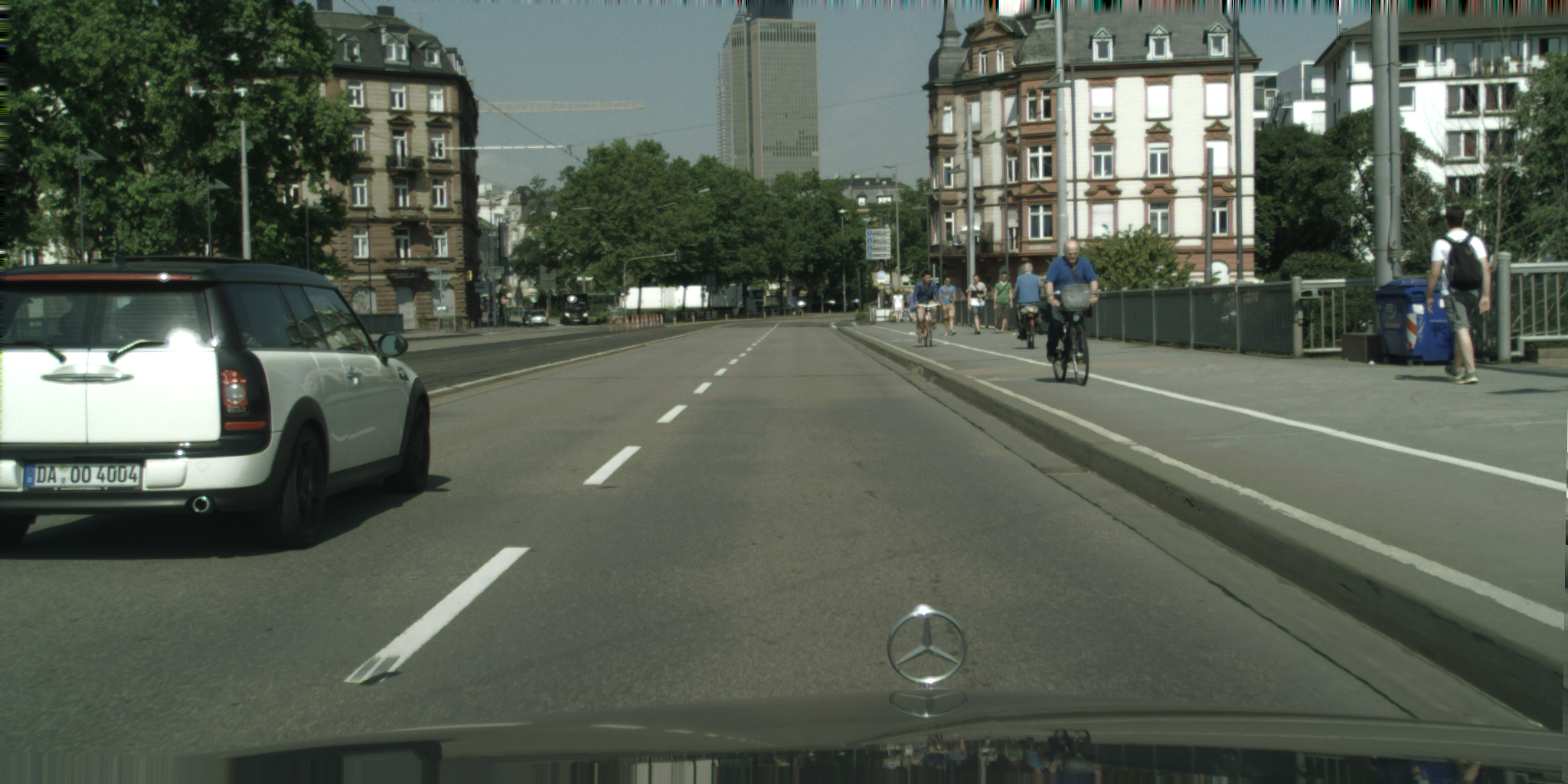} & \includegraphics[width=0.18\linewidth]{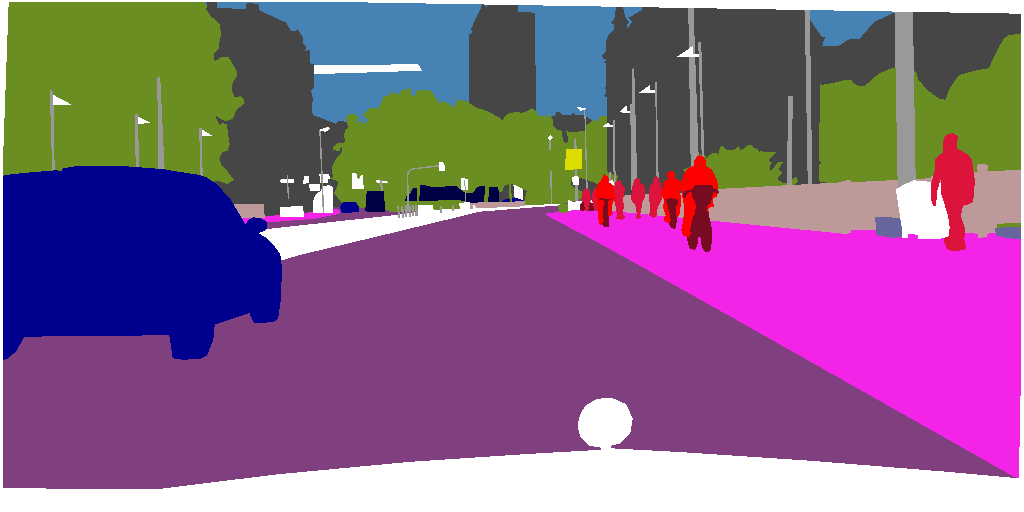} & \includegraphics[width=0.18\linewidth]{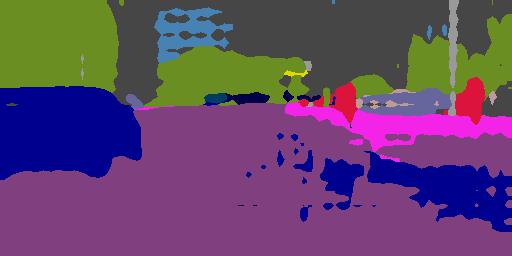} &
         \includegraphics[width=0.18\linewidth]{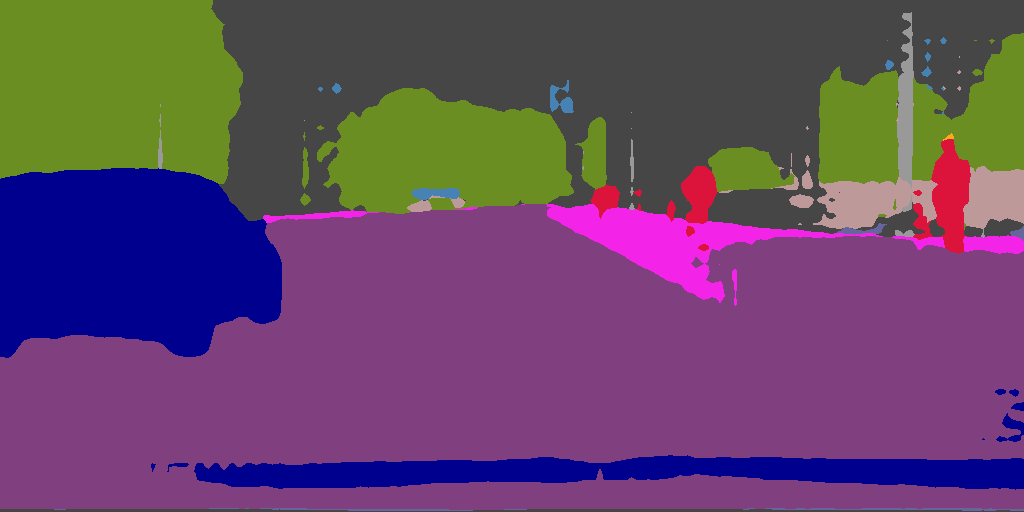} & \includegraphics[width=0.18\linewidth]{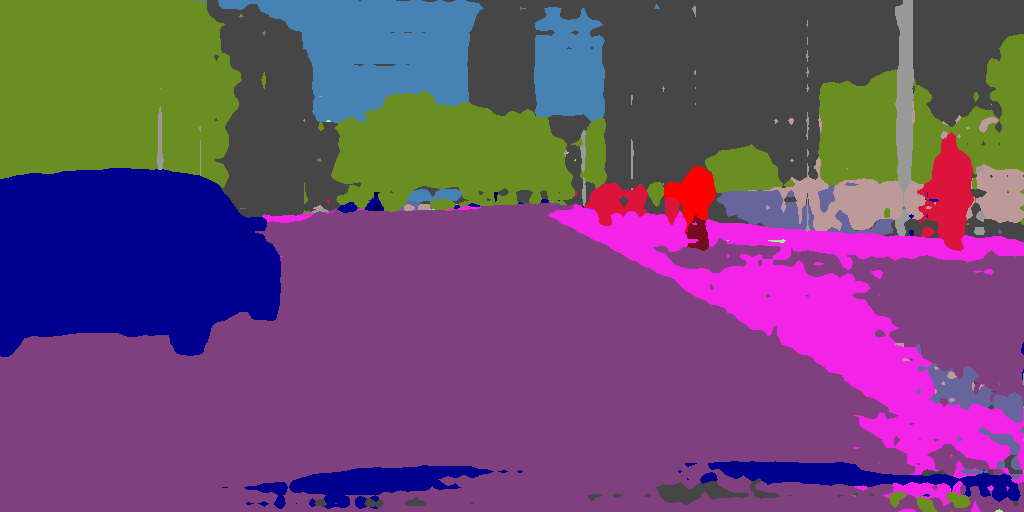} \\
         \includegraphics[width=0.18\linewidth]{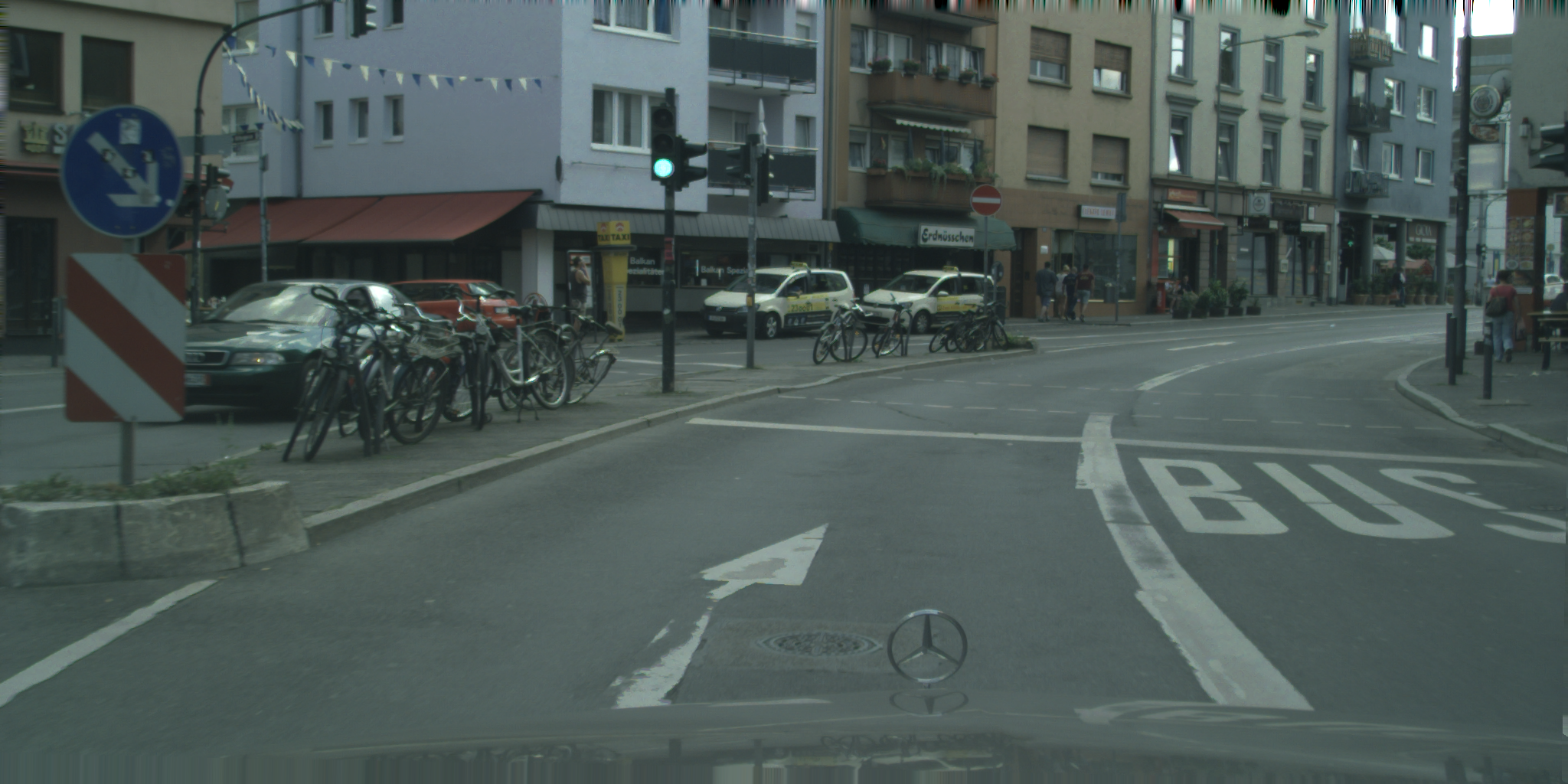} & \includegraphics[width=0.18\linewidth]{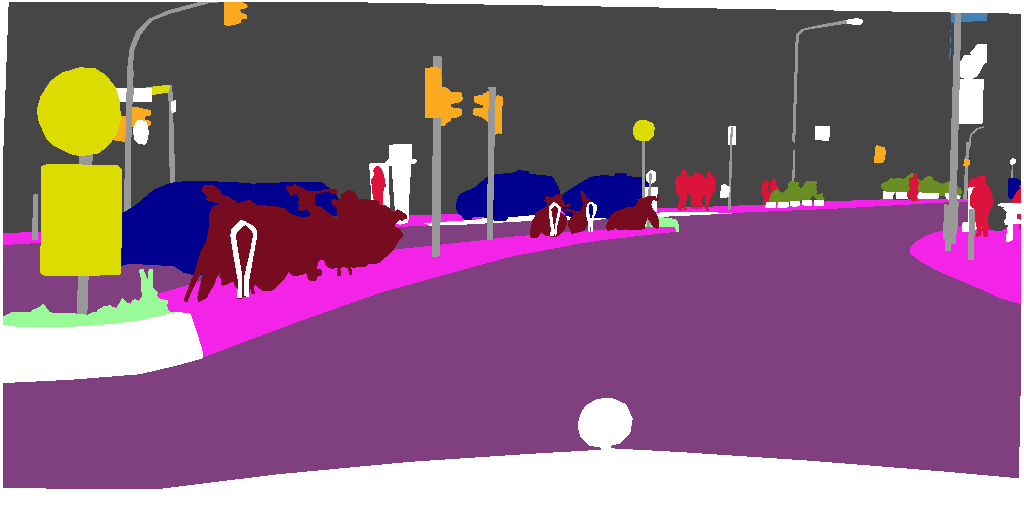} & \includegraphics[width=0.18\linewidth]{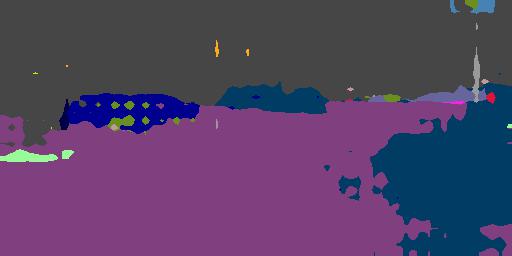} &
         \includegraphics[width=0.18\linewidth]{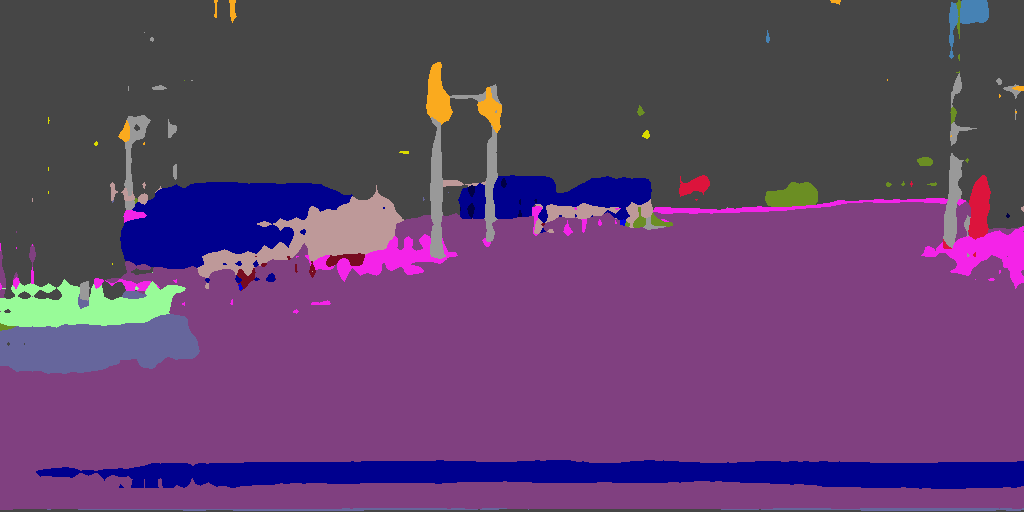} & \includegraphics[width=0.18\linewidth]{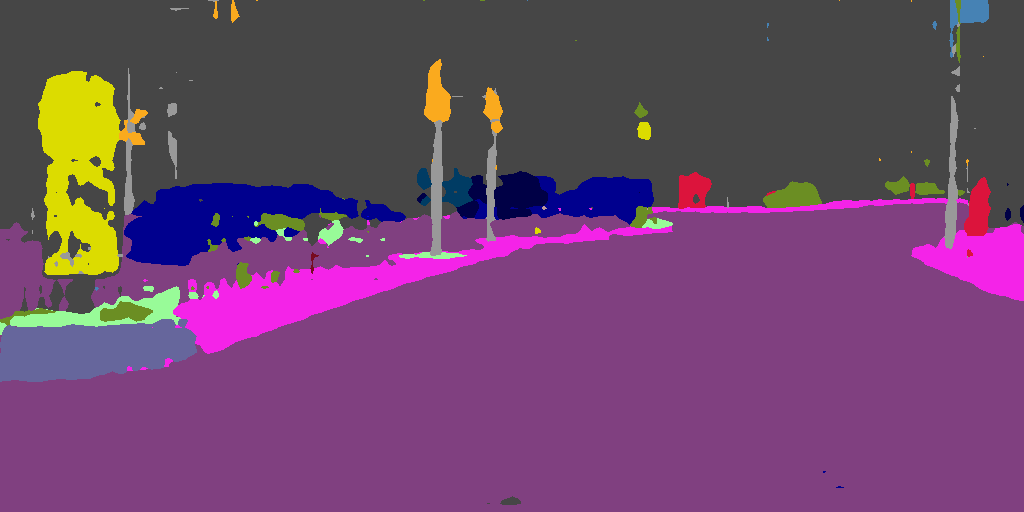} \\
         \includegraphics[width=0.18\linewidth]{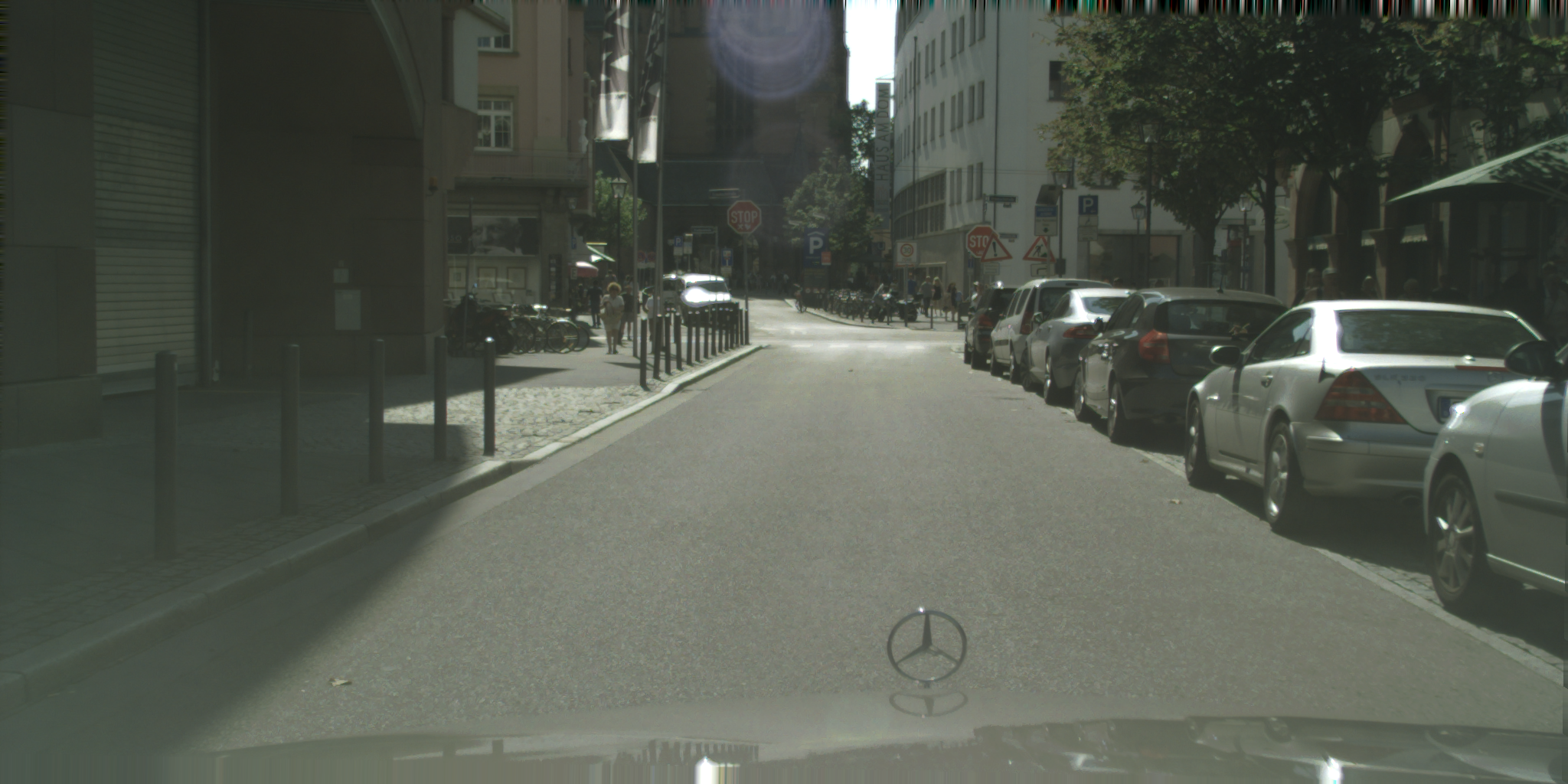} & \includegraphics[width=0.18\linewidth]{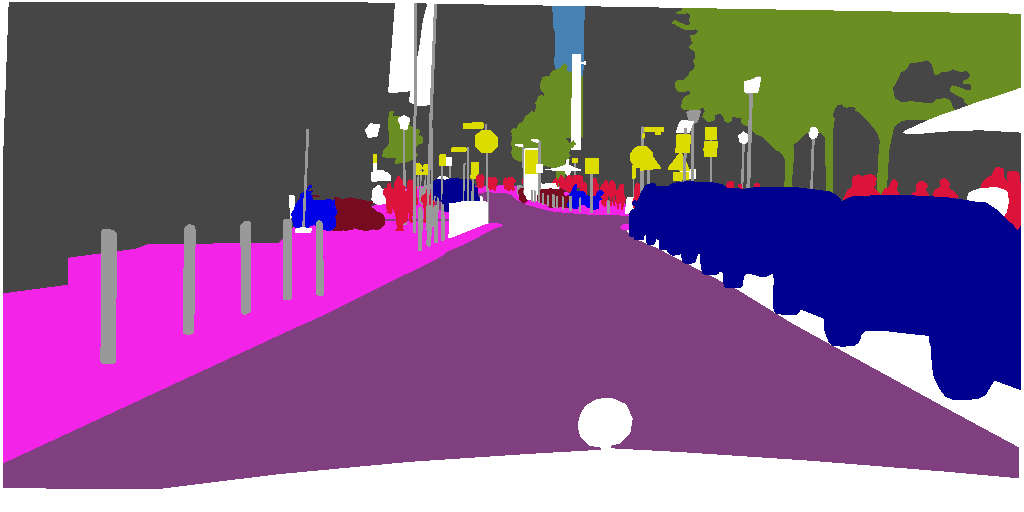} & \includegraphics[width=0.18\linewidth]{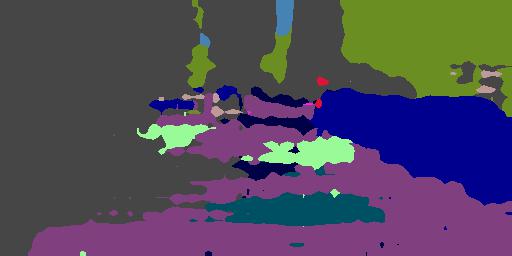} &
         \includegraphics[width=0.18\linewidth]{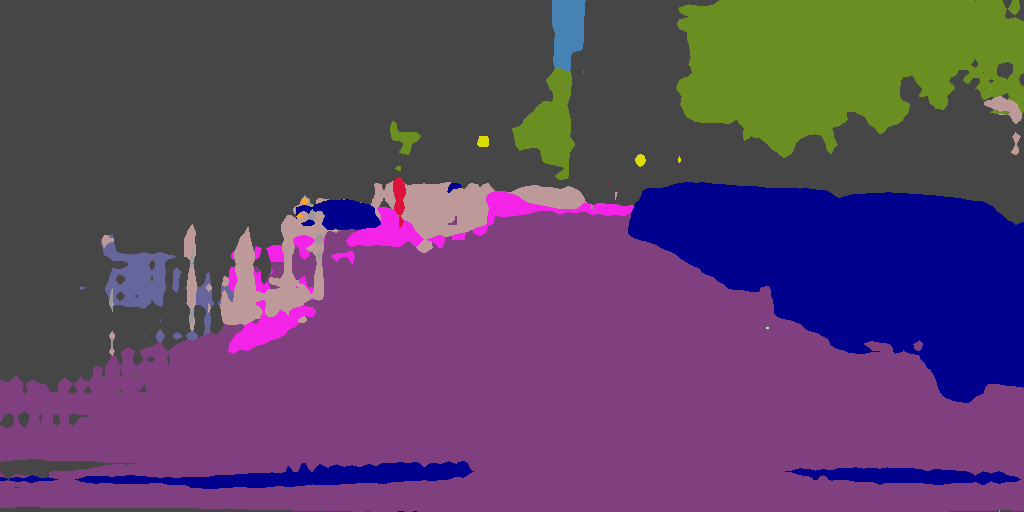} & \includegraphics[width=0.18\linewidth]{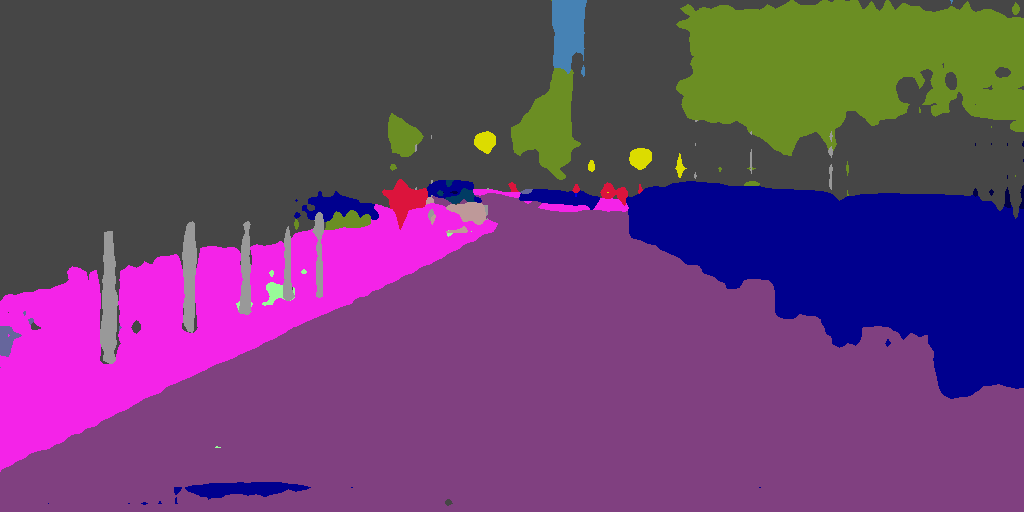} \\
         (a) Target Image & (b) Ground Truth & (c) Source Only &  (d) Conventional Adapt. & (e) DISE (ours)
    \end{tabular}
    \caption{Segmentation results on Cityscapes when adapted from GTA5. From left to right, (a) Target Image, (b) Ground Truth, (c) Source Only, (d) Conventional Adaptation~\cite{tsai2018learning}, (e) and DISE.}
    \label{fig:exp_result}
\end{figure*}

\subsection{Ablation Study}\label{ssec:exp_ablation}
The following presents a study of four variants of our model by comparing their performance with four distinct training objectives:
\begin{itemize}
    \item \textbf{Source Only}: Training with annotated GTA5 dataset~\cite{richter2016playing} by minimizing $\mathcal{L}_{seg}^s$ only, i.e. without any domain adaptation.
    
    \item \textbf{Seg-map Adaptation}: Training with annotated GTA5 dataset~\cite{richter2016playing} together with domain adaptation at the output space by minimizing $\mathcal{L}_{seg}^s$ and $\mathcal{L}_{seg\_adv}$. This corresponds to the method in~\cite{tsai2018learning}, which aligns segmentation predictions across domains. 
    
    
    \item \textbf{DISE w/o Label Transfer}: Training with all loss functions except label transfer loss, i.e. the setting for seg-map adaptation plus disentanglement of structure and texture components.  
    
    \item \textbf{DISE}: Training with all loss functions.  
        
\end{itemize}


Table \ref{table:ablation_loss} compares the performance of these settings in terms of mIoU. As expected, without any domain adaptation, "Source Only" shows the worst performance with a 39.8 mIoU. The performance improves by 2.8 with Seg-map Adaptation", arriving at a 42.6 mIoU, when introducing domain adaptation at the output space. An even higher gain of 4.3 over "Source Only" is seen for the setting of "DISE w/o Label Transfer", confirming the benefit of disentangling the structure and texture components. Finally, with additional augmented data due to label transfer, the DISE achieves the best performance.

\begin{table}[]
\caption{Ablation study results on Cityscapes when adapted from GTA5 in terms of mIoU. We present results for no adaptation (Source Only), adaptation at the output space only (Seg-map Adaptation), adaptation at the output space together with structure and texture disentanglement (DISE w/o Label Transfer), and adaptation with all losses considered (DISE).} \label{table:ablation_loss}
\centering
\begin{tabular}{lllllc}
\hline
\multicolumn{1}{l|}{Method} & \multicolumn{1}{c}{A} & \multicolumn{1}{c}{B} & \multicolumn{1}{c}{C} & \multicolumn{1}{c|}{D} & \multicolumn{1}{c}{mIoU} \\ \hline
\multicolumn{1}{l|}{Source Only} & \checkmark &  &  & \multicolumn{1}{l|}{} & 39.8 \\
\multicolumn{1}{l|}{Seg-map Adaptation} & \checkmark & \checkmark &  & \multicolumn{1}{l|}{} & 42.6 \\
\multicolumn{1}{l|}{DISE w/o Label Transfer} & \checkmark & \checkmark & \checkmark & \multicolumn{1}{l|}{} & 44.1 \\ 
\multicolumn{1}{l|}{DISE} & \checkmark & \checkmark & \checkmark & \multicolumn{1}{l|}{\checkmark} & 45.4 \\ \hline
\multicolumn{6}{l}{A: $\mathcal{L}_{seg}^s$} \\
\multicolumn{6}{l}{B: $\mathcal{L}_{seg\_adv}$} \\
\multicolumn{6}{l}{C: $\mathcal{L}_{rec}
	    +\mathcal{L}_{trans\_str} + \mathcal{L}_{trans\_tex}
	    + \mathcal{L}_{trans\_adv}$} \\
\multicolumn{6}{l}{D: $\mathcal{L}_{seg}^{s2t}$} \\ \hline
\end{tabular}
\end{table}
\subsection{Image-to-Image Translation}\label{ssec:exp_disentangle}
\begin{figure*}
    \centering
    \setlength\tabcolsep{0.25em}
    \begin{tabular}{@{}cc|cc|cc@{}}
         \multirow{6}{*}{\rot{S2T}} &
         \includegraphics[width=0.18\linewidth]{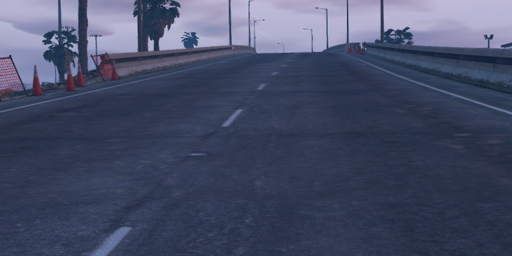} & \includegraphics[width=0.18\linewidth]{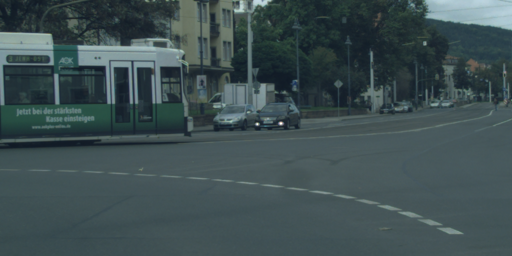} & \includegraphics[width=0.18\linewidth]{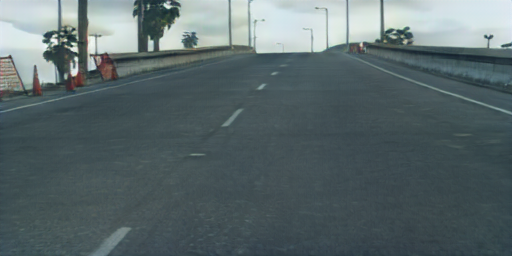} &
         \includegraphics[width=0.18\linewidth]{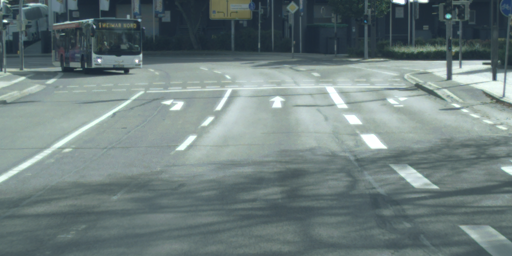} & \includegraphics[width=0.18\linewidth]{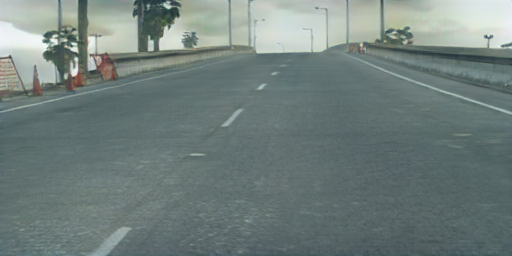} \\ 
         &
          \includegraphics[width=0.18\linewidth]{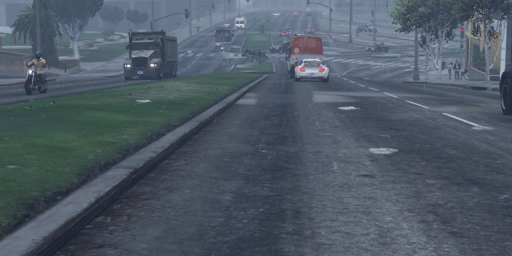} & \includegraphics[width=0.18\linewidth]{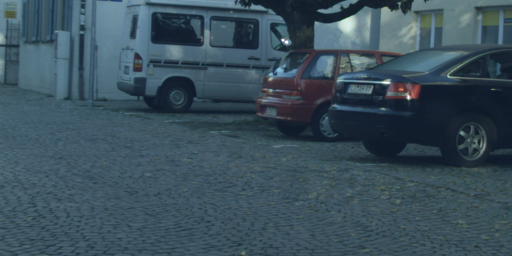} & \includegraphics[width=0.18\linewidth]{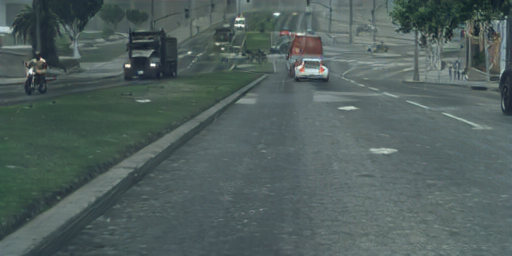} &
         \includegraphics[width=0.18\linewidth]{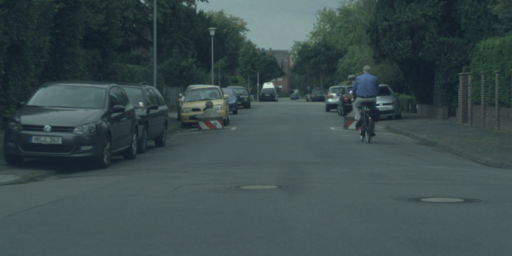} & \includegraphics[width=0.18\linewidth]{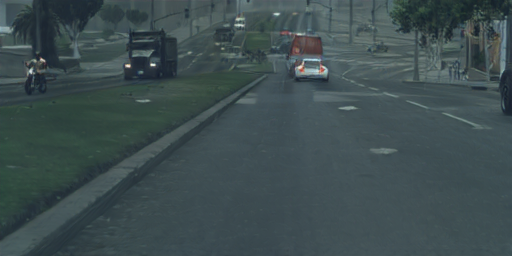} \\ 
         &
         \includegraphics[width=0.18\linewidth]{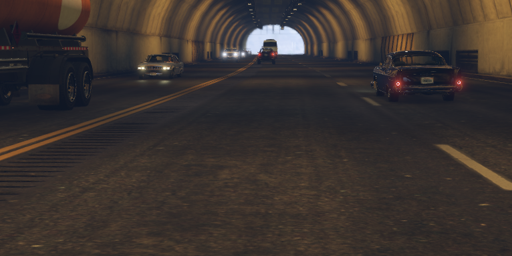} & \includegraphics[width=0.18\linewidth]{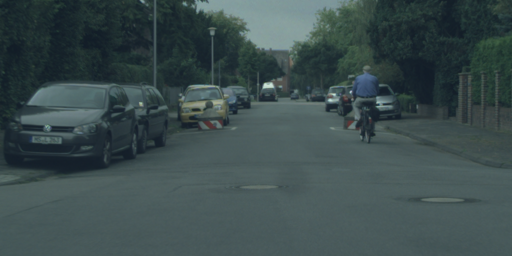} & \includegraphics[width=0.18\linewidth]{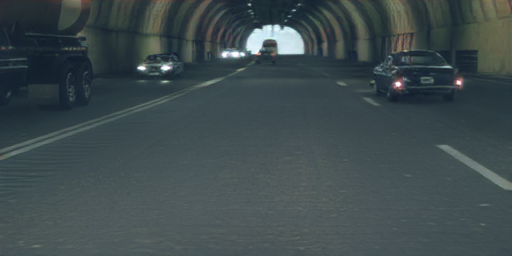} &
         \includegraphics[width=0.18\linewidth]{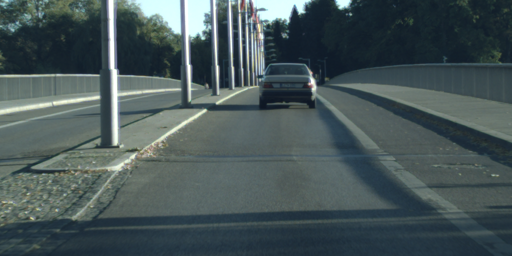} & \includegraphics[width=0.18\linewidth]{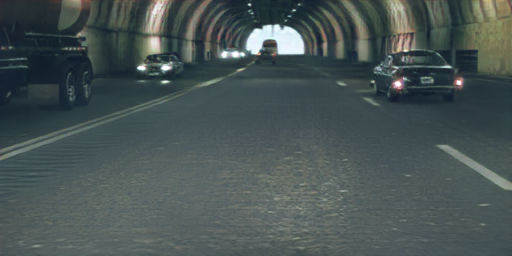}\\ \hline & & & &\\ [-8pt]
         \multirow{6}{*}{\rot{T2S}} &
         \includegraphics[width=0.18\linewidth]{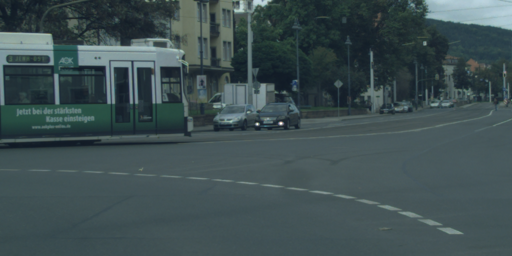} & \includegraphics[width=0.18\linewidth]{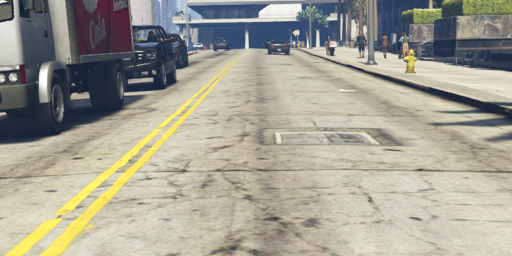} & \includegraphics[width=0.18\linewidth]{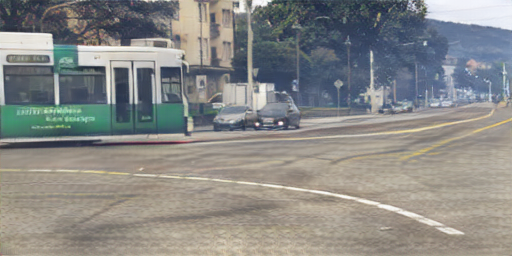} &
         \includegraphics[width=0.18\linewidth]{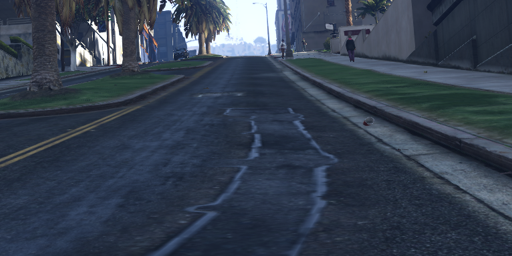} & \includegraphics[width=0.18\linewidth]{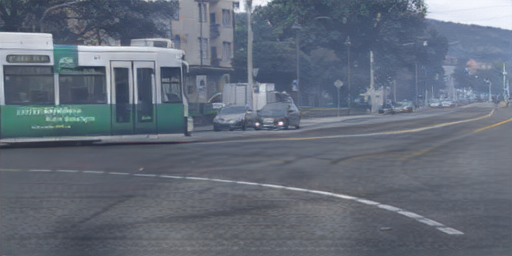} \\ 
         &
         \includegraphics[width=0.18\linewidth]{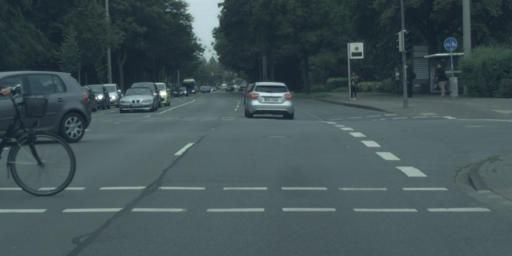} & \includegraphics[width=0.18\linewidth]{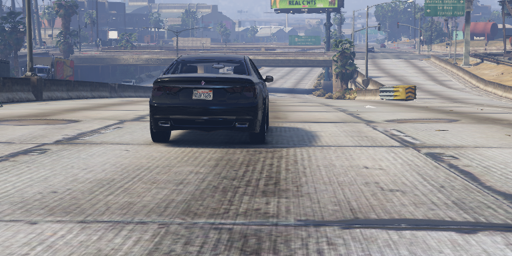} & \includegraphics[width=0.18\linewidth]{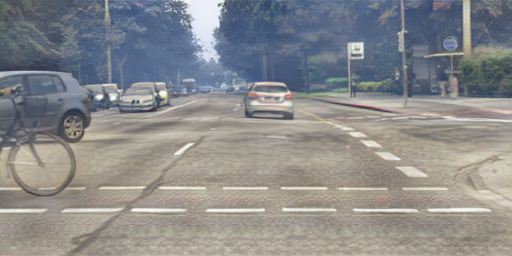} &
         \includegraphics[width=0.18\linewidth]{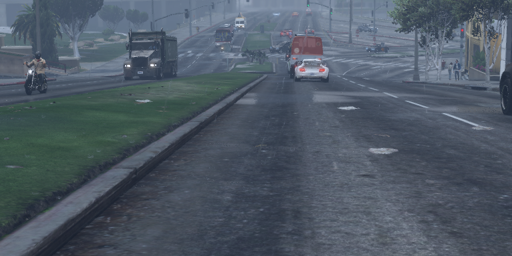} & \includegraphics[width=0.18\linewidth]{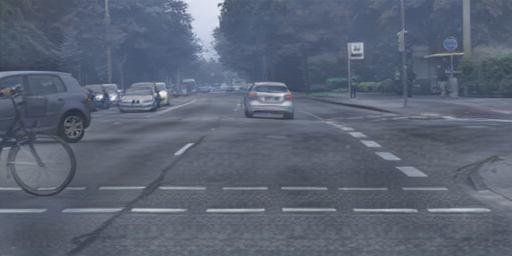} \\ 
         &
         \includegraphics[width=0.18\linewidth]{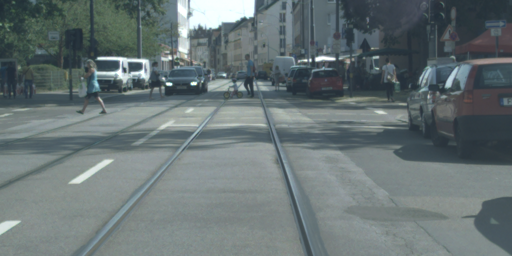} & \includegraphics[width=0.18\linewidth]{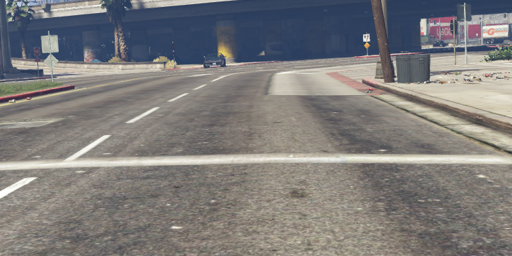} & \includegraphics[width=0.18\linewidth]{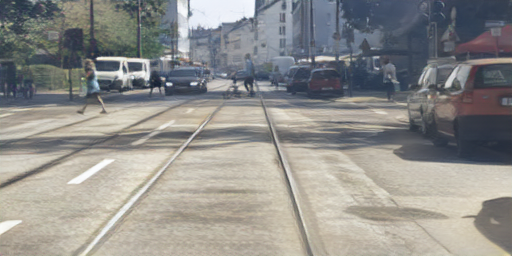} &
         \includegraphics[width=0.18\linewidth]{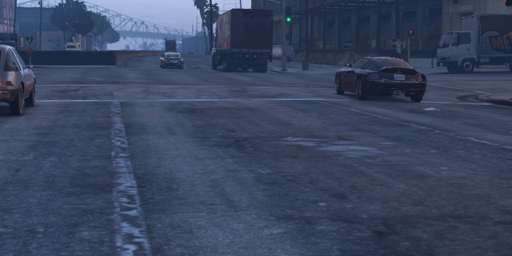} & \includegraphics[width=0.18\linewidth]{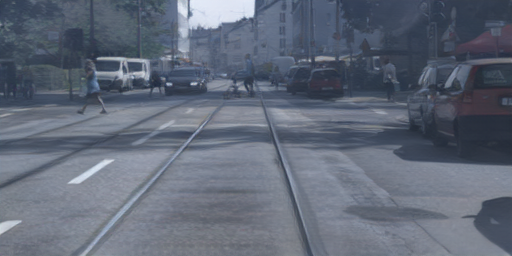} \\
         & (a) Structure & (b) Texture & (c) Output & (d) Texture & (e) Output \\
    \end{tabular}
    \caption{Sample results of translated images. S2T: the structure content of GTA5 images in (a) are combined with the texture appearance of Cityscapes images in (b) and (d) to output translated images in (c) and (e), respectively. T2S: the structure content of Cityscapes images in (a) are combined with the texture appearance of GTA5 images in (b) and (d) to output translated images in (c) and (e), respectively.}
    \label{fig:exp_disentangle}
\end{figure*}

In Figure \ref{fig:exp_disentangle}, we show qualitative results of image-to-image translation with DISE for two settings, S2T and T2S. With S2T (respectively, T2S), we combine the structure content of images in GTA5 (respectively, Cityscapes) in column (a) with the texture appearance of images in Cityscapes (respectively, GTA5) in columns (b) and (d) to produce translated images in columns (c) and (e), respectively. We see that DISE is very effective in translating images from one domain to another with high quality. In all cases, the translated images preserve well the structure content while producing the desired texture appearance. This also validates our use of the ground-truth labels of the source-domain images as pseudo labels for their translated images with texture appearance similar to target-domain images.    



\section{Conclusion}\label{sec:conclude}
In this paper, we hypothesize that the high-level structure information of an image is most decisive to semantic segmentation and can be made invariant across domains. Based on this hypothesis, we propose a novel framework, Domain Invariant Structure Extraction (DISE), to disentangle the representation of an image into a domain-invariant structure component and a domain-specific texture component, where the former is used to advance domain adaptation for semantic segmentation. The DISE also allows transfer of ground-truth labels from the source domain to the target domain, providing additional supervision for learning a segmentation network suitable for target-domain images. Extensive simulation results on typical datasets confirms the superiority of DISE over several state-of-the-art methods, justifying our initial hypothesis.

\section*{Acknowledgements}
\noindent This project is supported by MOST-108-2634-F-009-013 and MOST-108-2636-E-009-001 and we are grateful to the National Center for High-performance Computing for computer time and facilities.
{\small
\bibliographystyle{ieee}
\bibliography{egbib}

\begin{thebibliography}{10}\itemsep=-1pt

\bibitem{bousmalis2016domain}
K.~Bousmalis, G.~Trigeorgis, N.~Silberman, D.~Krishnan, and D.~Erhan.
\newblock Domain separation networks.
\newblock In {\em Advances in Neural Information Processing Systems (NIPS)},
  2016.

\bibitem{chen2018deeplab}
L.-C. Chen, G.~Papandreou, I.~Kokkinos, K.~Murphy, and A.~L. Yuille.
\newblock Deeplab: Semantic image segmentation with deep convolutional nets,
  atrous convolution, and fully connected crfs.
\newblock In {\em IEEE Transactions on Pattern Analysis and Machine
  Intelligence (TPAMI)}, 2018.

\bibitem{chen2018road}
Y.~Chen, W.~Li, and L.~Van~Gool.
\newblock Road: Reality oriented adaptation for semantic segmentation of urban
  scenes.
\newblock In {\em Proceedings of the IEEE Conference on Computer Vision and
  Pattern Recognition (CVPR)}, 2018.

\bibitem{cordts2016cityscapes}
M.~Cordts, M.~Omran, S.~Ramos, T.~Rehfeld, M.~Enzweiler, R.~Benenson,
  U.~Franke, S.~Roth, and B.~Schiele.
\newblock The cityscapes dataset for semantic urban scene understanding.
\newblock In {\em Proceedings of the IEEE Conference on Computer Vision and
  Pattern Recognition (CVPR)}, 2016.

\bibitem{everingham2015pascal}
M.~Everingham, S.~A. Eslami, L.~Van~Gool, C.~K. Williams, J.~Winn, and
  A.~Zisserman.
\newblock The pascal visual object classes challenge: A retrospective.
\newblock {\em International Journal of Computer Vision (IJCV)}, 2015.

\bibitem{ganin2014unsupervised}
Y.~Ganin and V.~Lempitsky.
\newblock Unsupervised domain adaptation by backpropagation.
\newblock In {\em Proceedings of the International Conference on Machine
  Learning (ICML)}, 2015.

\bibitem{he2016deep}
K.~He, X.~Zhang, S.~Ren, and J.~Sun.
\newblock Deep residual learning for image recognition.
\newblock In {\em Proceedings of the IEEE Conference on Computer Vision and
  Pattern Recognition (CVPR)}, 2016.

\bibitem{Hoffman_cycada2017}
J.~Hoffman, E.~Tzeng, T.~Park, J.-Y. Zhu, P.~Isola, K.~Saenko, A.~A. Efros, and
  T.~Darrell.
\newblock Cycada: Cycle consistent adversarial domain adaptation.
\newblock In {\em Proceedings of the International Conference on Machine
  Learning (ICML)}, 2018.

\bibitem{hong2018conditional}
W.~Hong, Z.~Wang, M.~Yang, and J.~Yuan.
\newblock Conditional generative adversarial network for structured domain
  adaptation.
\newblock In {\em Proceedings of the IEEE Conference on Computer Vision and
  Pattern Recognition (CVPR)}, 2018.

\bibitem{huang2017arbitrary}
X.~Huang and S.~J. Belongie.
\newblock Arbitrary style transfer in real-time with adaptive instance
  normalization.
\newblock In {\em Proceedings of the IEEE International Conference on Computer
  Vision (ICCV)}, 2017.

\bibitem{isola2017image}
P.~Isola, J.-Y. Zhu, T.~Zhou, and A.~A. Efros.
\newblock Image-to-image translation with conditional adversarial networks.
\newblock In {\em Proceedings of the IEEE Conference on Computer Vision and
  Pattern Recognition (CVPR)}, 2017.

\bibitem{Johnson2016Perceptual}
J.~Johnson, A.~Alahi, and L.~Fei-Fei.
\newblock Perceptual losses for real-time style transfer and super-resolution.
\newblock In {\em Proceedings of the European Conference on Computer Vision
  (ECCV)}, 2016.

\bibitem{lee2018diverse}
H.-Y. Lee, H.-Y. Tseng, J.-B. Huang, M.~Singh, and M.-H. Yang.
\newblock Diverse image-to-image translation via disentangled representations.
\newblock In {\em Proceedings of the European Conference on Computer Vision
  (ECCV)}, 2018.

\bibitem{liu2017unsupervised}
M.-Y. Liu, T.~Breuel, and J.~Kautz.
\newblock Unsupervised image-to-image translation networks.
\newblock In {\em Advances in Neural Information Processing Systems (NIPS)},
  2017.

\bibitem{long2015fully}
J.~Long, E.~Shelhamer, and T.~Darrell.
\newblock Fully convolutional networks for semantic segmentation.
\newblock In {\em Proceedings of the IEEE Conference on Computer Vision and
  Pattern Recognition (CVPR)}, 2015.

\bibitem{mao2017least}
X.~Mao, Q.~Li, H.~Xie, R.~Y. Lau, Z.~Wang, and S.~P. Smolley.
\newblock Least squares generative adversarial networks.
\newblock In {\em Proceedings of the IEEE International Conference on Computer
  Vision (ICCV)}, 2017.

\bibitem{richter2016playing}
S.~R. Richter, V.~Vineet, S.~Roth, and V.~Koltun.
\newblock Playing for data: Ground truth from computer games.
\newblock In {\em Proceedings of the European Conference on Computer Vision
  (ECCV)}, 2016.

\bibitem{ros2016synthia}
G.~Ros, L.~Sellart, J.~Materzynska, D.~Vazquez, and A.~M. Lopez.
\newblock The synthia dataset: A large collection of synthetic images for
  semantic segmentation of urban scenes.
\newblock In {\em Proceedings of the IEEE Conference on Computer Vision and
  Pattern Recognition (CVPR)}, 2016.

\bibitem{saleh2018effective}
F.~S. Saleh, M.~S. Aliakbarian, M.~Salzmann, L.~Petersson, and J.~M. Alvarez.
\newblock Effective use of synthetic data for urban scene semantic
  segmentation⋆.
\newblock In {\em Proceedings of the European Conference on Computer Vision
  (ECCV)}, 2018.

\bibitem{sankaranarayanan2018learning}
S.~Sankaranarayanan, Y.~Balaji, A.~Jain, S.~N. Lim, and R.~Chellappa.
\newblock Learning from synthetic data: Addressing domain shift for semantic
  segmentation.
\newblock In {\em Proceedings of the IEEE Conference on Computer Vision and
  Pattern Recognition (CVPR)}, 2018.

\bibitem{simonyan2014very}
K.~Simonyan and A.~Zisserman.
\newblock Very deep convolutional networks for large-scale image recognition.
\newblock In {\em Proceedings of the International Conference on Learning
  Representations (ICLR)}, 2014.

\bibitem{Simonyan14c}
K.~Simonyan and A.~Zisserman.
\newblock Very deep convolutional networks for large-scale image recognition.
\newblock {\em CoRR}, abs/1409.1556, 2014.

\bibitem{sun2016deep}
B.~Sun and K.~Saenko.
\newblock Deep coral: Correlation alignment for deep domain adaptation.
\newblock In {\em Proceedings of the European Conference on Computer Vision
  (ECCV)}, 2016.

\bibitem{tsai2018learning}
Y.-H. Tsai, W.-C. Hung, S.~Schulter, K.~Sohn, M.-H. Yang, and M.~Chandraker.
\newblock Learning to adapt structured output space for semantic segmentation.
\newblock In {\em Proceedings of the IEEE Conference on Computer Vision and
  Pattern Recognition (CVPR)}, 2018.

\bibitem{tzeng2017adversarial}
E.~Tzeng, J.~Hoffman, K.~Saenko, and T.~Darrell.
\newblock Adversarial discriminative domain adaptation.
\newblock In {\em Proceedings of the IEEE Conference on Computer Vision and
  Pattern Recognition (CVPR)}, 2017.

\bibitem{wu2018dcan}
Z.~Wu, X.~Han, Y.-L. Lin, M.~G. Uzunbas, T.~Goldstein, S.~N. Lim, and L.~S.
  Davis.
\newblock Dcan: Dual channel-wise alignment networks for unsupervised scene
  adaptation.
\newblock In {\em Proceedings of the European Conference on Computer Vision
  (ECCV)}, 2018.

\bibitem{yu2015multi}
F.~Yu and V.~Koltun.
\newblock Multi-scale context aggregation by dilated convolutions.
\newblock In {\em Proceedings of the International Conference on Learning
  Representations (ICLR)}, 2015.

\bibitem{zhao2017pyramid}
H.~Zhao, J.~Shi, X.~Qi, X.~Wang, and J.~Jia.
\newblock Pyramid scene parsing network.
\newblock In {\em Proceedings of the IEEE Conference on Computer Vision and
  Pattern Recognition (CVPR)}, 2017.

\bibitem{zhu2017unpaired}
J.-Y. Zhu, T.~Park, P.~Isola, and A.~A. Efros.
\newblock Unpaired image-to-image translation using cycle-consistent
  adversarial networks.
\newblock In {\em Proceedings of the IEEE Conference on Computer Vision and
  Pattern Recognition (CVPR)}, 2017.

\bibitem{zhu2018penalizing}
X.~Zhu, H.~Zhou, C.~Yang, J.~Shi, and D.~Lin.
\newblock Penalizing top performers: Conservative loss for semantic
  segmentation adaptation.
\newblock In {\em Proceedings of the European Conference on Computer Vision
  (ECCV)}, 2018.

\end{thebibliography}
}

\end{document}